\documentclass[runningheads]{llncs}

\usepackage{eccv}

\usepackage{eccvabbrv}

\usepackage{graphicx}
\usepackage{booktabs}

\usepackage[accsupp]{axessibility}  

\usepackage{hyperref}

\usepackage{orcidlink}

\usepackage{algorithm}
\usepackage{algorithmic}
\usepackage{amsmath} 
\usepackage{amssymb} 
\usepackage{array}
\usepackage{booktabs}
\usepackage{multirow}
\usepackage{adjustbox}
\usepackage[table]{xcolor}
\usepackage{caption}
\usepackage{tcolorbox}
\tcbuselibrary{breakable} 
\usepackage{enumitem}     
\usepackage{marvosym}
\begin{document}

\title{EventMemAgent: Hierarchical Event-Centric Memory for Online Video Understanding with Adaptive Tool Use} 

\titlerunning{EventMemAgent}

\author{
    Siwei Wen\inst{1} \and 
    Zhangcheng Wang\inst{3} \and 
    Xingjian Zhang\inst{1} \and 
    Lei Huang\inst{1,2}\,\textsuperscript{\Letter} \and
    Wenjun Wu\inst{1,2}
}
\authorrunning{S. Wen et al.}

\institute{
    Beijing Advanced Innovation Center for Future Blockchain and Privacy Computing, School of Artificial Intelligence, Beihang University, Beijing, China \and
    Hangzhou International Innovation Institute, Beihang University, Hangzhou, China \and 
    4Paradigm Inc., Beijing, China\\
    \email{huangleiai@buaa.edu.cn}
}

\maketitle
\begin{abstract}

Online video understanding requires models to perform continuous perception and long-range reasoning within potentially infinite visual streams. Its fundamental challenge lies in the conflict between the unbounded nature of streaming media input and the limited context window of Multimodal Large Language Models (MLLMs). Current methods primarily rely on passive processing, which often face a trade-off between maintaining long-range context and capturing the fine-grained details necessary for complex tasks. To address this, we introduce \textbf{EventMemAgent}, an active online video agent framework based on a \textbf{hierarchical memory module}. Our framework employs a dual-layer strategy for online videos: short-term memory detects event boundaries and utilizes event-granular reservoir sampling to process streaming video frames within a fixed-length buffer dynamically; long-term memory structuredly archives past observations on an event-by-event basis. Furthermore, we integrate a \textbf{multi-granular perception toolkit} for active, iterative evidence capture and employ \textbf{Agentic Reinforcement Learning} (Agentic RL) to end-to-end internalize reasoning and tool-use strategies into the agent's intrinsic capabilities.  Experiments show that \textbf{EventMemAgent} achieves competitive results on online video benchmarks. The code is available at \url{https://github.com/lingcco/EventMemAgent}.
\keywords{Online video understanding \and Video agent}
\end{abstract}
\section{Introduction}
\label{sec:intro}

Online video understanding has become a pivotal task in numerous real-world applications, such as autonomous driving, intelligent surveillance, and robotic assistants. Unlike offline video analysis, these scenarios require models to perform continuous perception and long-range reasoning within potentially infinite visual streams under constrained computational and memory budgets. The fundamental challenge lies in the conflict between the unbounded nature of streaming inputs and the finite context window of MLLMs.

Current studies have primarily sought to alleviate this challenge by managing historical information through passive processing. One line of work focuses on visual token management within the context window, employing techniques such as token pruning or sliding windows \cite{xu2025streamingvlm, qian2024streaming, zheng2025videoscaffold, zeng2025streamforest, zhang2025flash}. While these methods can mitigate the issue, they inevitably suffer from the decay and loss of historical information as the video stream progresses, and they remain structurally limited by the maximum context window. Another research direction utilizes external memory to preserve visual tokens from past streaming video \cite{chen2025streamkv, di2025streaming, zhang2026hermes, yang2025streamagent, ye2025venus, guan2026video}. During inference, visual information with high similarity to the question is manually added to the model context as a reference. However, such storage imposes substantial overhead and lacks the semantic interpretability required for complex reasoning. More importantly, passive perceiving methods often struggle to achieve precise comprehension of fine-grained details when processing a vast number of visual tokens in a single pass.

We argue that overcoming these limitations requires a paradigm shift from a passive information processing paradigm to an active agentic framework. Instead of merely storing and recalling information passively, an agent can decompose complex video understanding tasks into multiple sub-tasks and actively, iteratively retrieve or perceive task-relevant information \cite{zuo2025videolucy, wang2024videoagent, ma2025drvideo, yang2025longvt, li2025perceive}. This approach allows the model to maintain long-range video understanding while executing fine-grained tasks with precision. However, the potential of such agentic intelligence in online video scenarios remains underexplored. Current online video agents are often restricted to limited toolsets and simplistically designed memory modules \cite{long2025seeing}, which constrains their effectiveness in online video understanding tasks. Furthermore, these systems rely heavily on manual prompt engineering \cite{yang2025streamagent}, failing to internalize reasoning and tool-invocation strategies into the agent's intrinsic capabilities through end-to-end training.

To bridge the gap between the active agentic paradigm and online video practice, we propose \textbf{EventMemAgent}, an online video agent framework based on a \textbf{hierarchical memory module}. Our framework introduces a dual-layer memory management strategy tailored for online video scenarios. Specifically, the short-term memory utilizes event-granular reservoir sampling to dynamically process streaming frames and detect event boundaries, while the long-term memory structuredly archives past observations into event-centric tuples, including captions, visual anchors, and change logs. Furthermore, we integrate a \textbf{multi-granular toolkit}, such as memory search and object detection, enabling the agent to actively and iteratively capture evidence across different levels of information granularity. Crucially, we employ \textbf{Agentic RL} to end-to-end optimize reasoning paths and tool-invocation policies, effectively internalizing the agentic framework into its intrinsic decision-making process.

The main contributions of this work are summarized as follows:

\begin{itemize}

    \item We propose \textbf{EventMemAgent}, an active online video agent framework that enables precise comprehension by decomposing tasks and iteratively perceiving relevant information. By utilizing Agentic RL, our framework internalizes reasoning and tool-use strategies into its intrinsic capabilities.
    
    \item We design a \textbf{hierarchical memory module} that coordinates a short-term memory for online event segmentation and event-granular reservoir sampling with a long-term memory for structured event-centric tuples. This dual-layer architecture enables efficient event-centric management of infinite video streams.

    \item Experiments show that \textbf{EventMemAgent} achieves strong results on online video benchmarks using lightweight visual input. This demonstrates that our framework can handle infinite video streams efficiently while staying within fixed hardware constraints.

\end{itemize}

\section{Related Work}
\label{sec:related_work}

\subsection{Online Video Understanding}

Online video understanding is designed to process continuous and potentially infinite visual inputs under fixed hardware constraints. However, most existing MLLMs \cite{liu2023visual, li2024llava, lin2024video, zhang2025tinyllava, zhang2025tinyllavar1, bai2025qwen2, bai2025qwen3vltechnicalreport} are primarily designed for predefined offline videos and struggle to handle online videos.

Therefore, recent research has focused primarily on processing online video through passive information processing strategies. One major research direction manages visual tokens through passive processing strategies within the context window, such as token pruning, sliding windows, and KV cache compression \cite{xu2025streamingvlm, qian2024streaming, zheng2025videoscaffold, zeng2025streamforest, zhang2025flash}.
For instance, \textit{StreamingVLM} \cite{xu2025streamingvlm} employs a sliding window to discard past information, while \textit{VideoStreaming} \cite{qian2024streaming} utilizes a trained encoder to compress historical information into fixed-length tokens.
While efficient, these passive methods inevitably suffer from information decay as the video stream progresses and remain structurally limited by the maximum context window of the underlying model.

Another research direction explores utilizing external memory modules to archive historical visual tokens \cite{chen2025streamkv, di2025streaming, zhang2026hermes, yang2025streamagent, ye2025venus}, manually retrieving relevant tokens based on similarity during inference.
For example, \textit{StreamKV} \cite{chen2025streamkv} compresses and offloads visual KV caches to external storage, while \textit{Venus} \cite{ye2025venus} selects keyframes and stores their embeddings in memory. Although these approaches circumvent context length constraints, their passive recall mechanisms hinder the model's ability to achieve precise comprehension among vast tokens. Furthermore, such feature-level storage lacks the semantic interpretability required for complex reasoning.

In contrast, \textbf{EventMemAgent} transitions from passive processing to an active agentic framework. Supported by an event-centric hierarchical memory, it ensures long-term coherence through dynamic segmentation and structured archiving. Crucially, our agent actively and iteratively perceives task-relevant evidence via a multi-granular toolkit, overcoming the limitations of passive similarity-based recall.

\subsection{Video Understanding Agents}

Recent advancements in long video understanding have explored agent-based frameworks as a viable paradigm for processing large-scale visual inputs \cite{zuo2025videolucy, wang2024videoagent, ma2025drvideo, yang2025longvt, li2025perceive, zhi2025videoagent2}. Unlike conventional video MLLMs that struggle with lengthy sequences, these systems leverage the high-level reasoning, planning, and memory management capabilities of Large Language Models (LLMs). By decomposing complex video tasks into manageable sub-tasks and iteratively retrieving and synthesizing task-relevant information, these agents effectively enhance the performance in long video understanding.

In the context of online video, although some recent works \cite{long2025seeing, yang2025streamagent} have begun to incorporate agentic frameworks, significant limitations persist. For instance, \textit{M3-Agent} \cite{long2025seeing} is restricted to rudimentary memory retrieval tools during inference, failing to leverage a diverse toolset to unlock the potential of the agentic paradigm; moreover, its simple text-only memory module, which relies on fixed-length segmentation, remains suboptimal for capturing long-term temporal dynamics. Meanwhile, \textit{StreamAgent} \cite{yang2025streamagent} relies heavily on manual prompt engineering to guide its decision-making, lacking exploration into utilizing Agentic RL to internalize reasoning and tool-invocation strategies as intrinsic capabilities. Our work addresses these limitations by introducing a hierarchical memory module and a multi-granular toolkit, leveraging reinforcement learning to optimize the agent's planning and execution within dynamic streaming environments.
\section{Methodology}

\subsection{Overview}
\begin{figure*}[!ht]
        \centering
        \includegraphics[width=1\linewidth]{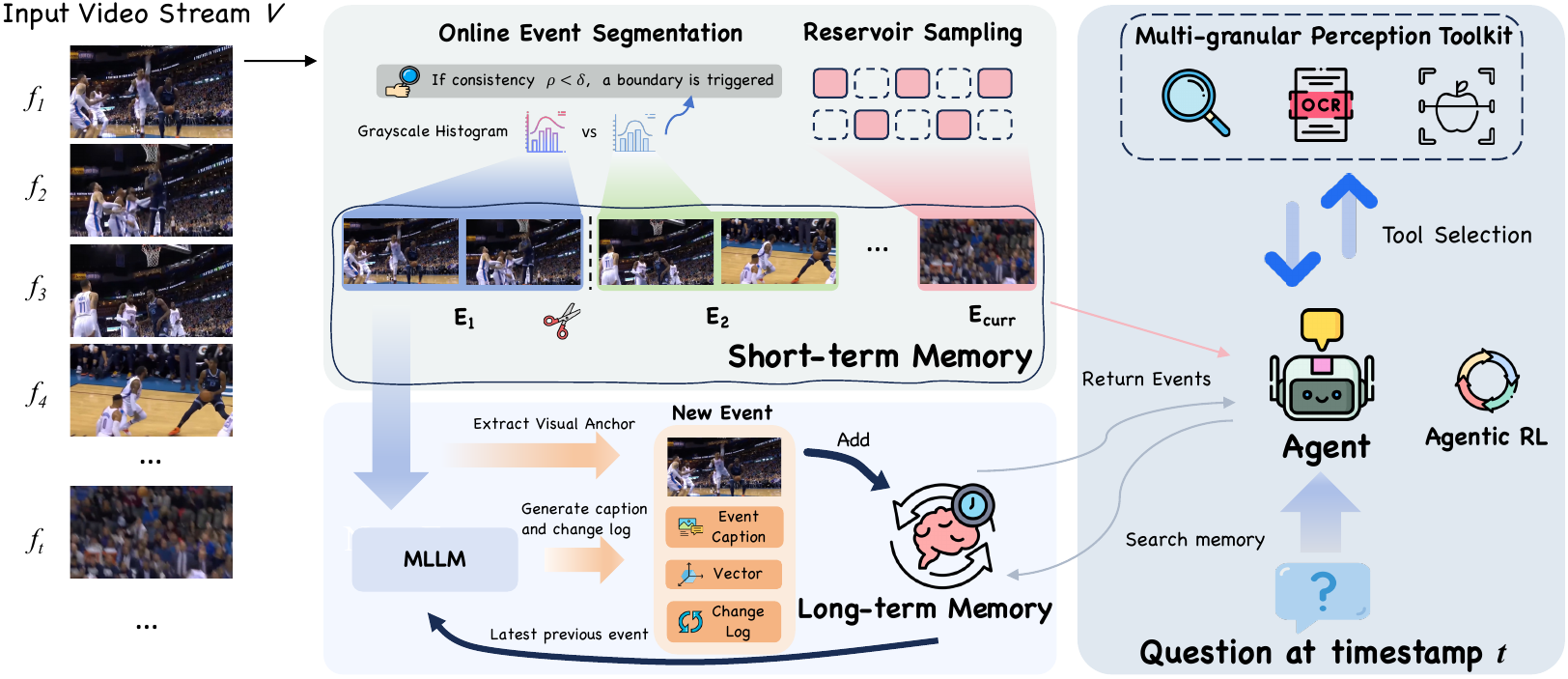}
        \caption{\textbf{Overview of EventMemAgent.} It consists of three core components: a hierarchical memory module that archives video streams into structured event-centric representations, a multi-granular perception toolkit for active, iterative evidence capture, and an agentic reinforcement learning framework that optimizes tool-use strategies. The agent dynamically retrieves memory and utilizes perception tools to answer questions at specific timestamps.}
        \label{fig:framework}
    \end{figure*}

We propose an active online video agent framework designed to shift the passive information processing paradigm toward proactive perception for online video. As illustrated in Figure \ref{fig:framework}, our architecture consists of three integrated components: (1) a \textbf{Hierarchical Memory Module} designed to overcome information decay and semantic fragmentation by dynamically archiving visual streams into structured, event-centric representations; (2) a \textbf{Multi-granular Perception Toolkit} that enables the agent to actively and iteratively capture fine-grained evidence across diverse levels of granularity; and (3) an \textbf{Agentic Reinforcement Learning} framework that internalizes reasoning and tool-invocation strategies into the agent's intrinsic capabilities through end-to-end optimization.

\subsection{Hierarchical Memory Module} 
To address the challenge of processing infinite video streams within a finite context window while overcoming information decay and semantic fragmentation, we design an event-centric hierarchical memory system. The architecture consists of a Short-Term Memory (STM) for immediate visual buffering and a Long-Term Memory (LTM) for structured history archiving. We consistently employ a sampling rate of 1 FPS on the incoming video stream $V = \{f_1, f_2, \dots, f_t\, \dots\}$, where $f_t$ denotes the frame captured at time step $t$, to extract discrete frames before proceeding with further processing.

\subsubsection{Event-Centric Short-Term Memory}

The Short-Term Memory operates as a fixed-size buffer with a maximum capacity of $K$ frames, directly accessible during inference. Instead of a flat frame sequence, we structure the STM as a collection of $m$ semantically coherent events. Formally, let $\mathcal{E}_{st}$ denote the set of events in the short-term memory:
\begin{equation}
\mathcal{E}_{st} = \{E_1, E_2, \dots, E_m\},
\end{equation}

\noindent where $\{E_1, \dots, E_{m-1}\}$ represents completed historical segments, and $E_m$ denotes the active event accumulating incoming frames. Each event $E_i = \{f_{i,j}\}_{j=1}^{n_i}$ consists of $n_i$ frames belonging to the same semantic segment. The total frame count is strictly constrained by $\sum_{i=1}^{m} n_i \le K$.

\paragraph{Online Event Segmentation.}
To maintain semantic coherence, for each incoming frame $f_t$, the system determines whether it belongs to the current active event $E_m$ or signifies a new segment. 
We evaluate content consistency by comparing the normalized grayscale histogram $\mathbf{h}_t$ of the new frame with the average histogram $\mathbf{\bar{h}}_m$ of all frames currently in $E_m$. 
Specifically, we employ the Pearson correlation coefficient $\rho$ as a quantitative metric to measure the distributional similarity between the new frame and the event history.
The coefficient is computed as:
\begin{equation}
\rho(\mathbf{\bar{h}}_m, \mathbf{h}_t) = \frac{\text{cov}(\mathbf{\bar{h}}_m, \mathbf{h}_t)}{\sigma_{\mathbf{\bar{h}}_m} \sigma_{\mathbf{h}_t}},
\end{equation}

\noindent where $\text{cov}(\cdot)$ and $\sigma$ denote the covariance and standard deviation calculated over the histogram bins, respectively. A boundary is triggered if the correlation $\rho < \delta$. Upon a boundary, the finalized $E_m$ is archived into the historical memory, and a new event $E_{m+1}$ is initialized with $f_t$.

\begin{figure}[t]
        \centering
        \includegraphics[width=1\linewidth]{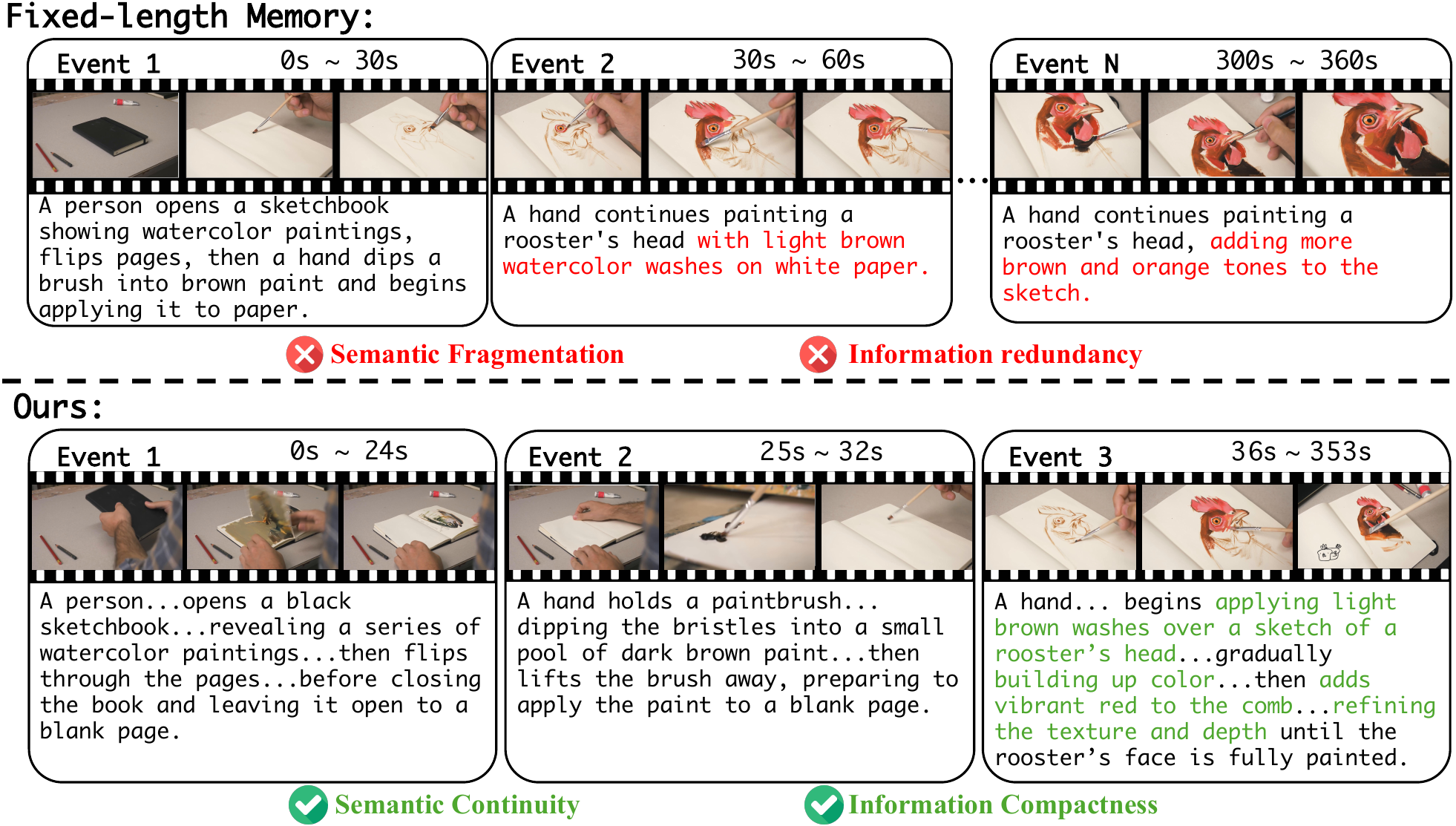}
        \caption{\textbf{Comparison of memory management strategies.} Fixed-length Memory (top) frequently suffers from semantic fragmentation and information redundancy due to rigid temporal boundaries that bisect continuous actions. In contrast, our event-centric hierarchical memory (bottom) preserves semantic continuity and information compactness.}
        \label{fig:memory}
    \end{figure}

\paragraph{Intra-event Reservoir Sampling.} 

Under our online event segmentation mechanism, semantically coherent events—such as those where the subject remains stationary—typically manifest substantial informational redundancy in the visual stream (As shown in Figure \ref{fig:memory}). To circumvent the inefficient accumulation of redundant data and preclude a single protracted event from depleting the $K$-frame context budget, we adopt a reservoir sampling strategy within $E_{m}$ when event segmentation is not triggered.

Specifically, incoming frames are directly appended to the active event $E_m$ as long as the total frame count across all STM events remains below $K$. Once the STM reaches its $K$-frame capacity, we resolve the overflow via two mutually exclusive branches:
(1) Inter-event FIFO: If the STM contains multiple events, we execute a First-In-First-Out strategy, transitioning the earliest event $E_1$ to the LTM to accommodate the new frame.
(2) Intra-event Reservoir Sampling: If $E_m$ is the sole event in the STM, indicating it has entirely saturated the global budget, we apply reservoir sampling. The $n$-th processed frame of this event ($n > K$) is accepted with probability $P = K/n$, replacing a randomly selected frame in $E_m$.

This mechanism guarantees that the inclusion probability for each frame remains uniform, ensuring that $E_{m}$ consistently serves as an unbiased representative summary of an event stream with an indeterminate duration. Consequently, this approach effectively maximizes the temporal horizon covered within the fixed context window while mitigating short-term memory overflow. See Appendix A for detailed short-term memory update rules.

\subsubsection{Structured Long-Term Memory}

When an event $E_i$ is evicted from the short-term buffer to accommodate new observations, it is archived in the Long-Term Memory as a structured tuple $E_i' = \{I_i^{first}, c_i, \mathbf{e}_i, \Delta_i\}$. Each entry is anchored by the first frame $I_i^{first}$, providing a visual reference, while the semantic content is captured by a natural language caption $c_i$ generated by MLLMs. To facilitate efficient retrieval, $c_i$ is projected into a high-dimensional space as a semantic embedding $\mathbf{e}_i$. Furthermore, we address the potential for narrative fragmentation by incorporating Change Logs $\Delta_i$, which record the state transitions and logical shifts between consecutive events. This hierarchical transition ensures that while the STM focus remains on immediate dynamics, the historical context is preserved indefinitely in a structured and searchable format.

In summary, the hierarchical memory module reconciles the inherent conflict between unbounded video input and limited context length. Furthermore, as illustrated in Figure \ref{fig:memory}, our event-centric design maximizes the utility of the limited short-term memory, enhancing narrative continuity and informational compactness while avoiding semantic fragmentation and redundancy common in fixed-length partitioning. Additionally, our long-term memory archives both captions and visual anchors to preserve more fine-grained details. Detailed visualizations of the memory contents are provided in Appendix E.

\subsection{Multi-granular Perception Toolkit}
To extend the agent's perceptual depth beyond raw visual signals, we integrate a toolkit $\mathcal{T}$ that enables active information acquisition. A key component is the \textit{Memory Search} tool, which empowers the agent to retrieve historical contexts from the LTM through two modalities: (i) Temporal Retrieval, where the agent filters events $E_i'$ whose timestamps $[t_i^{start}, t_i^{end}]$ intersect with a specific query range; and (ii) Semantic Retrieval, which identifies relevant episodes by maximizing the cosine similarity between the query embedding $\mathbf{e}_q$ and stored event embeddings $\mathbf{e}_i$. The retrieval returns the complete event content from the LTM, such as the caption and change log.

Beyond memory retrieval, the agent invokes specialized perception tools to capture fine-grained evidence. Specifically, it utilizes \textit{OCR} to extract textual cues and \textit{Object Detection} to localize specific entities. The agent can autonomously choose to apply these operations either to visual anchors associated with events in the LTM or to specific frames within the STM. The observations are integrated into the context to facilitate further reasoning and response generation.

\subsection{Agentic Reinforcement Learning}

The agent follows a multi-turn reasoning trajectory adhering to the ReAct paradigm \cite{yao2022react}:
\begin{equation}
\tau = \{(t_1, a_1, o_1), \dots, (t_n, a_n, o_n)\},
\end{equation}

\noindent In this structure, \textit{Thought} ($t$) represents the agent's internal reasoning for task decomposition, \textit{Action} ($a$) specifies either a tool invocation or the terminal response, and \textit{Observation} ($o$) captures the sensory feedback returned by the perception toolkit. 

To effectively internalize the agentic framework's reasoning and tool-invocation strategies into its intrinsic decision-making capabilities, we employ Group Relative Policy Optimization (GRPO) \cite{guo2025deepseek}. For each query $q$, we sample a group of trajectories $\{\tau_1, \dots, \tau_G\}$ from the old policy $\pi_{\theta_{old}}$. GRPO optimizes the following objective by utilizing group-based relative advantages instead of a separate critic:

\begin{equation}
\resizebox{0.9\linewidth}{!}{$\displaystyle 
\mathcal{J}_{GRPO}(\theta) = \mathbb{E} \left[ \frac{1}{G} \sum_{i=1}^G \min \left( \frac{\pi_\theta(\tau_i|q)}{\pi_{\theta_{old}}(\tau_i|q)} A_i, \text{clip} \left( \frac{\pi_\theta(\tau_i|q)}{\pi_{\theta_{old}}(\tau_i|q)}, 1-\epsilon, 1+\epsilon \right) A_i \right) \right]
$}
\end{equation}

\noindent where the advantage $A_i$ is computed by normalizing the rewards within each group: $A_i = (r_i - \text{mean}(\mathbf{r})) / \text{std}(\mathbf{r})$. 

In our framework, the reward is assigned solely based on the terminal correctness of the final answer:
\begin{equation}
r_i = 
\begin{cases} 
1, & \text{if the predicted answer is correct} \\
0, & \text{otherwise}
\end{cases}
.
\end{equation}

\section{Experiments}
\subsection{Experimental Setup}

\begin{table*}[!b]
  \centering
  \caption{The experimental results on the OVO-Bench benchmark. The benchmark contains many tasks, including Optical Character Recognition (OCR), Action Recognition (ACR), Attribute Recognition (ATR), Spatial Understanding (STU), Future Prediction (FPD), Object Recognition (OJR), Episodic Memory (EPM), Action Sequence Identification (ASI), Hallucination Detection (HLD), Repetition Event Count (REC), Sequential Steps Recognition (SSR), and Clues Reveal Responding (CRR).}
  \renewcommand{\arraystretch}{1.7}
  \setlength{\tabcolsep}{1pt}
  \vspace{-0.6em}
  \resizebox{\linewidth}{!}{
      \begin{tabular}{lcc|ccccccc|cccc|cccc|c}
        \hline
        \multirow{2}{*}{Model} & \multirow{2}{*}{Params} & \multirow{2}{*}{Frames} & \multicolumn{7}{c|}{Real-Time Visual Perception}  & \multicolumn{4}{c|}{Backward Tracing} & \multicolumn{4}{c|}{Forward Active Responding} & \multirow{2}{*}{\textbf{Overall}} \\
        & & & OCR & ACR & ATR & STU & FPD & OJR & \textbf{Avg.} & EPM & ASI & HLD & \textbf{Avg.} & REC & SSR & CRR & \textbf{Avg.} &  \\
        \hline
        \multicolumn{19}{c}{\textbf{Human}}\\
        \hline
        Human Agents & - & - & 93.96 & 92.57 & 94.83 & 92.70 & 91.09 & 94.02 & 93.20 & 92.59 & 93.02 & 91.37 & 92.33 & 95.48 & 89.67 & 93.56 & 92.90 & 92.81 \\
        \hline
        \multicolumn{19}{c}{\textbf{Proprietary MLLMs}}\\
        \hline
        Gemini 1.5 Pro~\cite{team2024gemini} & - & 1fps & 85.91 & 66.97 & 79.31 & 58.43 & 63.37 & 61.96 & 69.32 & 58.59 & 76.35 & 52.64 & 62.54 & 35.53 & 74.24 & 61.67 & 57.15 & 63.00 \\
        GPT-4o~\cite{hurst2024gpt} & - & 64 & 69.8 & 64.22 & 71.55 & 51.12 & 70.3 & 59.78 & 64.46 & 57.91 & 75.68 & 48.66 & 60.75 & 27.58 & 73.21 & 59.4 & 53.40 & 59.54 \\
        \hline
        \multicolumn{19}{c}{\textbf{Open-Source Offline MLLMs}}\\
        \hline
        Qwen2-VL-72B~\cite{wang2024qwen2} & 72B & 64 & 65.77 & 60.55 & 69.83 & 51.69 & 69.31 & 54.35 & 61.92 & 52.53 & 60.81 & 57.53 & 56.95 & 38.83 & 64.07 & 45.00 & 49.30 & 56.27 \\
        LLaVA-Video-7B~\cite{lin2024video} & 7B & 64 & 69.13 & 58.72 & 68.83 & 49.44 & 74.26 & 59.78 & 63.52 & 56.23 & 57.43 & 7.53 & 40.4 & 34.10 & 69.95 & 60.42 & 54.82 & 52.91 \\
        LLaVA-OneVision-7B~\cite{li2024llava} & 7B & 64 & 66.44 & 57.80 & 73.28 & 53.37 & 71.29 & 61.96 & 64.02 & 54.21 & 55.41 & 21.51 & 43.71 & 25.64 & 67.09 & 58.75 & 50.50 & 52.74 \\
        Qwen2-VL-7B~\cite{wang2024qwen2} & 7B & 64 & 60.40 & 50.46 & 56.03 & 47.19 & 66.34 & 55.43 & 55.98 & 47.81 & 35.48 & 56.08 & 46.46 & 31.66 & 65.82 & 48.75 & 48.74 & 50.39 \\
        InternVL-V2-8B~\cite{chen2024far} & 8B & 64 & 67.11 & 60.55 & 63.79 & 46.07 & 68.32 & 56.52 & 60.39 & 48.15 & 57.43 & 24.73 & 43.44 & 26.5 & 59.14 & 54.14 & 46.60 & 50.15 \\
        LongVU-7B~\cite{shen2024longvu} & 7B & 1fps & 53.69 & 53.21 & 62.93 & 47.75 & 68.32 & 59.78 & 57.61 & 40.74 & 59.46 & 4.84 & 35.01 & 12.18 & 69.48 & 60.83 & 47.50 & 46.71 \\
        Qwen3-VL-8B~\cite{bai2025qwen3vltechnicalreport} & 8B & 64 & 75.17 & 58.72 & 72.41 & 57.30 & 70.30 & 59.24 & 65.52 & 56.57 & 69.59 & 12.37 & 46.18 & 47.13 & 67.57 & 52.50 & 55.73 & 55.81 \\
        \hline
        \multicolumn{19}{c}{\textbf{Open-Source Online MLLMs}}\\
        \hline
        Flash-VStream-7B~\cite{zhang2025flash} & 7B & 1fps & 24.16 & 29.36 & 28.45 & 33.71 & 25.74 & 28.80 & 28.37 & 39.06 & 37.16 & 5.91 & 27.38 & 8.02 & 67.25 & 60.00 & 45.09 & 33.61 \\
        VideoLLM-online-8B~\cite{chen2024videollm} & 8B & 2fps & 8.05 & 23.85 & 12.07 & 14.04 & 45.54 & 21.20 & 20.79 & 22.22 & 18.80 & 12.18 & 17.73 & - & - & - & - & - \\
        Dispider~\cite{qian2025dispider} & 7B & 1fps & 57.72 & 49.54 & 62.07 & 44.94 & 61.39 & 51.63 & 54.55 & 48.48 & 55.41 & 4.3 & 36.06 & 18.05 & 37.36 & 48.75 & 34.72 & 41.78 \\
        StreamForest-7B~\cite{zeng2025streamforest} & 7B & 1fps & 68.46 & 53.21 & 71.55 & 47.75 & 65.35 & 60.87 & 61.20 & 58.92 & 64.86 & 32.26 & 52.02 & 32.81 & 70.59 & 57.08 & 53.49 & 55.57 \\
        \hline
        \multicolumn{19}{c}{\textbf{Online Video Agents}}\\
        \hline
        StreamAgent~\cite{yang2025streamagent} & 7B & 1fps & 71.20 & 53.20 & 63.60 & 53.90 & 67.30 & 58.70 & 61.30 & 54.80 & 58.10 & 25.80 & 41.70 & 35.90 & 48.40 & 52.00 & 45.40 & 49.40 \\
        Ours & 8B & $\le$32 & 75.84 & 69.72 & 73.28 & 55.62 & 67.33 & 67.93 & 68.29 & 59.60 & 70.95 & 43.55 & 58.03 & 33.67 & 72.02 & 62.08 & 55.92 & 60.75 \\
        \hline
      \end{tabular}
    }
    \vspace{3mm}
  \label{tab:ovobench_result}
\end{table*}

\paragraph{Benchmarks.} We evaluate \textbf{EventMemAgent} on two popular benchmarks for online video question answering: OVO-Bench \cite{niu2025ovo} and StreamingBench \cite{lin2024streamingbench}. 
In these benchmarks, models must follow the online video format, which means they can only use the video content before the question's timestamp to answer. 
OVO-Bench divides online video tasks into three scenarios: 
(1) Real-Time Visual Perception, 
(2) Backward Tracing, and 
(3) Forward Active Responding, 
to fully test model performance in different online video settings. 
StreamingBench includes 12 tasks to evaluate real-time understanding in various situations.


\paragraph{Implementation Details.} We employ Qwen3-VL-8B-\allowbreak Instruct \cite{bai2025qwen3vltechnicalreport} as the backbone MLLM for \textbf{EventMemAgent}, using Grounding DINO \cite{liu2024grounding} as the object detection model and Deepseek-OCR \cite{wei2025deepseek} as the OCR model. By default, input videos are sampled at 1 FPS, and the maximum capacity of short-term memory is set to 32. The content consistency threshold $\delta$ is set to 0.2. In short-term memory, we added frame numbers and timestamps to the images to improve the model's spatiotemporal perception \cite{wu2025number}.  Since Qwen3-VL already has basic tool-use capabilities, we use 10K MovieChat \cite{song2024moviechat} samples labeled by VideoMarathon \cite{lin2025unleashing} for direct end-to-end agentic RL training. All training and evaluation are conducted on eight 80G A100 GPUs.
For more deployment details, please refer to Appendix B.

\paragraph{Compared Baselines.}
Our compared baselines mainly include: 
(1) Proprietary MLLMs, such as GPT-4o \cite{hurst2024gpt} and Gemini-1.5-Pro \cite{team2024gemini}. 
(2) Advanced open-source offline MLLMs, including InternVL-V2 \cite{chen2024far}, Qwen3-VL \cite{bai2025qwen3vltechnicalreport}, and LLaVA-Video \cite{lin2024video}. To adapt these offline models to the online video setting, fixed FPS or uniform sampling is applied to the video segment before the query timestamp.
(3) Recent open-source online MLLMs, such as Flash-VStream \cite{zhang2024flash} and Dispider \cite{qian2025dispider}. (4) Online video agents, StreamAgent \cite{yang2025streamagent}.

\subsection{Main Results}


\paragraph{Results on OVO-Bench.}  We report the detailed results of \textbf{EventMemAgent} on OVO-Bench in Table~\ref{tab:ovobench_result}. With input frames limited to 32, \textbf{EventMemAgent} achieves an average accuracy of 60.75\%, surpassing all existing open-source models and even the proprietary model \textbf{GPT-4o} (59.54\%). Specifically, our model outperforms current open-source methods by 4.27\% in Real-Time Visual Perception, 1.08\% in Backward Tracing, and 1.1\% in Forward Active Responding. These gains are driven by our hierarchical event-centric memory, which organizes streaming video into structured event-level representations. Furthermore, the multi-granular perception tools and agentic RL training allow the model to actively capture fine-grained details and internalize efficient tool-use strategies. This combination ensures high-fidelity perception and precise reasoning within a limited context, leading to superior stability and accuracy across diverse online video scenarios.

\begin{table*}[!t]
    \centering
    \renewcommand{\arraystretch}{1}
    \caption{The experimental results on the real-time visual understanding part of the StreamingBench benchmark. The benchmark contains many tasks, including Object Perception (OP), Causal Reasoning (CR), Clips Summarization (CS), Attribute Perception (ATP), Event Understanding (EU), Text-Rich Understanding (TR), Prospective Reasoning (PR), Spatial Understanding (SU), Action Perception (ACP), and Counting (CT).}
    \label{tab:streamingbench}
    \small
    \begin{adjustbox}{max width=\linewidth}
    \begin{tabular}{l c c | c c c c c c c c c c | c}
    \toprule
    \multirow{1}{*}{Model} & Params & \multirow{1}{*}{Frames} & OP & CR & CS & ATP & EU & TR & PR & SU & ACP & CT & \textbf{ALL} \\
    \midrule
    Human & - & - & 89.47 & 92.00 & 93.60 & 91.47 & 95.65 & 92.52 & 88.00 & 88.75 & 89.74 & 91.30 & 91.46 \\
    \midrule
    \multicolumn{14}{c}{\textbf{Proprietary MLLMs}} \\
    \midrule
    Gemini 1.5 pro~\cite{team2024gemini} & - & 1 fps & 79.02 & 80.47 & 83.54 & 79.67 & 80.00 & 84.74 & 77.78 & 64.23 & 71.95 & 48.70 & 75.69 \\
    GPT-4o~\cite{hurst2024gpt} & - & 64  
        & 77.11 & 80.47 & 83.91 & 76.47 & 70.19 & 83.80 & 66.67 & 62.19 & 69.12 & 49.22 & 73.28 \\
    Claude 3.5 Sonnet~\cite{Claude3S} & - & 20  
        & 73.33 & 80.47 & 84.09 & 82.02 & 75.39 & 79.53 & 61.11 & 61.79 & 69.32 & 43.09 & 72.44 \\
    \midrule
    \multicolumn{14}{c}{\textbf{Open-source Offline MLLMs}} \\
    \midrule
    Video-LLaMA2~\cite{cheng2024videollama} & 7B & 32 & 
            55.86 & 55.47 & 57.41 & 58.17 & 52.80 & 43.61 & 39.81 & 42.68 & 45.61 & 35.23 & 49.52 \\
    VILA-1.5~\cite{liu2025nvila} & 8B & 14 &  
        53.68 & 49.22 & 70.98 & 56.86 & 53.42 & 53.89 & 54.63 & 48.78 & 50.14 & 17.62 & 52.32 \\
    Video-CCAM~\cite{fei2024video} & 14B & 96 & 
        56.40 & 57.81 & 65.30 & 62.75 & 64.60 & 51.40 & 42.59 & 47.97 & 49.58 & 31.61 & 53.96 \\
    LongVA~\cite{zhang2024long} & 7B & 128 & 
        70.03 & 63.28 & 61.20 & 70.92 & 62.73 & 59.50 & 61.11 & 53.66 & 54.67 & 34.72 & 59.96 \\
    InternVL-V2~\cite{chen2024far} & 8B & 16 & 
        68.12 & 60.94 & 69.40 & 77.12 & 67.70 & 62.93 & 59.26 & 53.25&  54.96 & 56.48 & 63.72 \\
    Qwen2-VL~\cite{wang2024qwen2} & 7B & 0.2-1fps & 75.20 & 77.45 & 73.19 & 82.81 & 68.32 & 71.03 & 72.22 & 61.19 & 61.47 & 46.11 & 69.04 \\
    Kangaroo~\cite{liu2024kangaroo} & 7B & 64 & 
        71.12 & 84.38 & 70.66 & 73.20 & 67.08 & 61.68 & 56.48 & 55.69 & 62.04 & 38.86 & 64.60 \\
    LLaVA-NeXT-Video~\cite{zhang2024llavanextvideo} & 32B & 64 & 
        78.20 & 70.31 & 73.82 & 76.80 & 63.35 & 69.78 & 57.41 & 56.10&  64.31 & 38.86 & 66.96 \\
    MiniCPM-V-2.6~\cite{yao2024minicpm} & 8B & 32 &  
        71.93 & 71.09 & 77.92 & 75.82 & 64.60 & 65.73 & 70.37 & 56.10 & 62.32 & 53.37 & 67.44 \\
    Qwen3-VL~\cite{bai2025qwen3vltechnicalreport} & 8B & 0.2-1 fps &
        73.57 & 71.09 & 76.34 & 80.39 & 65.22 & 74.77 & 78.70 & 67.89 & 61.47 & 47.67 & 70.20 \\
    \addlinespace[1pt]
     
    \midrule
    \multicolumn{14}{c}{\textbf{Open-source Online MLLMs}} \\
    \midrule
    Flash-VStream~\cite{zhang2024flash} & 7B & -  
        & 25.89 & 43.57 & 24.91 & 23.87 & 27.33 & 13.08 & 18.52 & 25.20 & 23.87 & 48.70 & 23.23 \\
    VideoLLM-online~\cite{chen2024videollm} & 8B & 2 fps  
        & 39.07 & 40.06 & 34.49 & 31.05 & 45.96 & 32.40 & 31.48 & 34.16 & 42.49 & 27.89 & 35.99 \\
    Dispider~\cite{qian2025dispider} & 7B & 1 fps  
        & 74.92 & 75.53 & 74.10 & 73.08 & 74.44 & 59.92 & 76.14 & 62.91 & 62.16 & 45.80 & 67.63 \\
    TimeChat-Online~\cite{yao2025timechat} & 7B & 1 fps
        & 80.22 & 82.03 & 79.50 & 83.33 & 76.10 & 78.50 & 78.70 & 64.63 & 69.60 & 57.98 & 75.36 \\
    StreamForest~\cite{zeng2025streamforest} & 7B & 1 fps
        & 83.11 & 82.81 & 82.65 & 84.26 & 77.50 & 78.19 & 76.85 & 69.11 & 75.64 & 54.40 & 77.26 \\
    \midrule
    \multicolumn{14}{c}{\textbf{Online Video Agents}} \\
    \midrule
    StreamAgent~\cite{yang2025streamagent} & 7B & 1fps & 79.63 & 78.31 & 79.28 & 75.87 & 74.74 & 76.92 & 82.94 & 66.31 & 73.69 & 55.40 & 74.28 \\
    Ours & 8B & $\le$32
        & 83.92 & 76.56 & 87.70 & 85.29 & 77.02 & 79.44 & 77.78 & 68.29 & 72.24 & 48.70 & 77.00 \\
    \bottomrule
    \end{tabular}
    \end{adjustbox}
\end{table*}

\paragraph{Results on StreamingBench.} We report the detailed performance of \textbf{EventMemAgent} on StreamingBench in Table~\ref{tab:streamingbench}, which covers 12 diverse real-time understanding tasks. Our model delivers highly competitive performance, demonstrating its strength in streaming video scenarios. A key observation is that success in these tasks depends heavily on the precise perception of current content rather than the complex integration of long-term history, which suggests that the potential of LTM in our framework has not yet been fully realized. Even with no more than 32 input frames, \textbf{EventMemAgent} achieves impressive accuracy, proving its efficiency in processing immediate visual streams. This capability is supported by our multi-granular perception tools and event-centric short-term memory, which enable the agent to capture essential details within the current event window.

\begin{figure*}[t!]
    \centering
    \includegraphics[width=0.95\linewidth]{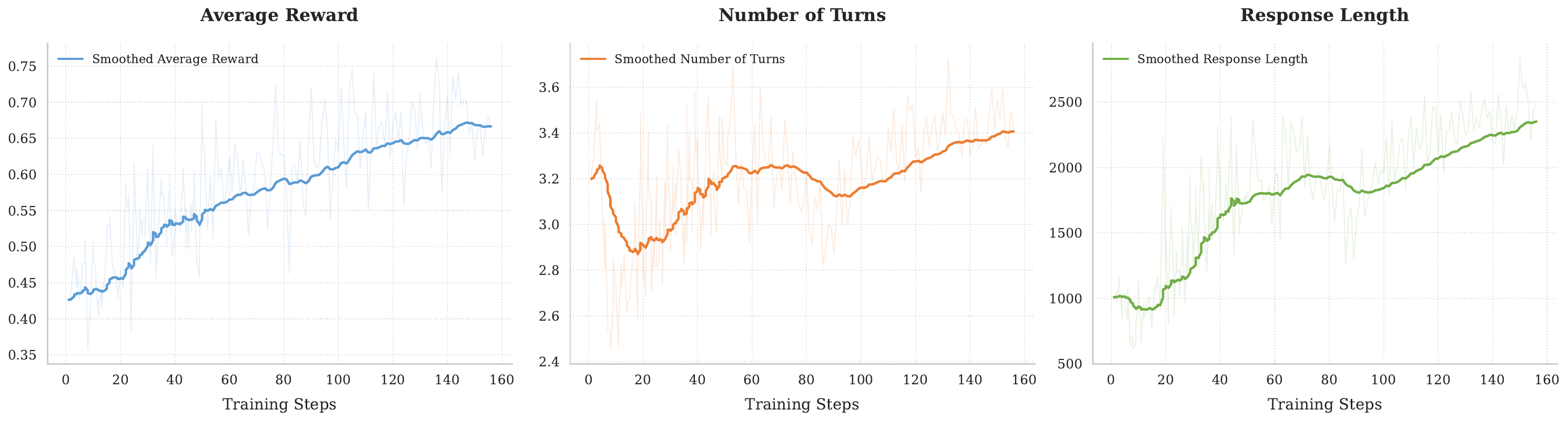}
    \caption{\textbf{Training Statistics of Agentic RL.} 
    The evolution of Average Reward (left), Number of Turns (middle), and Response Length (right) throughout the training process. 
    The steady increase in both reward and response length demonstrates the model's improving capability in complex reasoning and tool usage.}
    \label{fig:training_curves}
\end{figure*}

\paragraph{Training Statistics} We visualize the training dynamics of the Agentic RL process in Figure~\ref{fig:training_curves}.
The \textbf{Average Reward} (left) exhibits a steady upward trend, confirming that the GRPO algorithm effectively optimizes the agent's policy.
Simultaneously, we observe a significant increase in the \textbf{Response Length} (right), which indicates that the model is learning to generate more detailed reasoning chains (CoT) and tool invocations to ensure answer correctness.
The \textbf{Number of Turns} (middle) initially fluctuates before stabilizing, reflecting the agent's adaptation to the multi-turn interaction environment.
These metrics collectively demonstrate the stability and effectiveness of our training framework.

\subsection{Efficiency Analysis}

\begin{figure*}[t!]
    \centering
    \includegraphics[width=\linewidth]{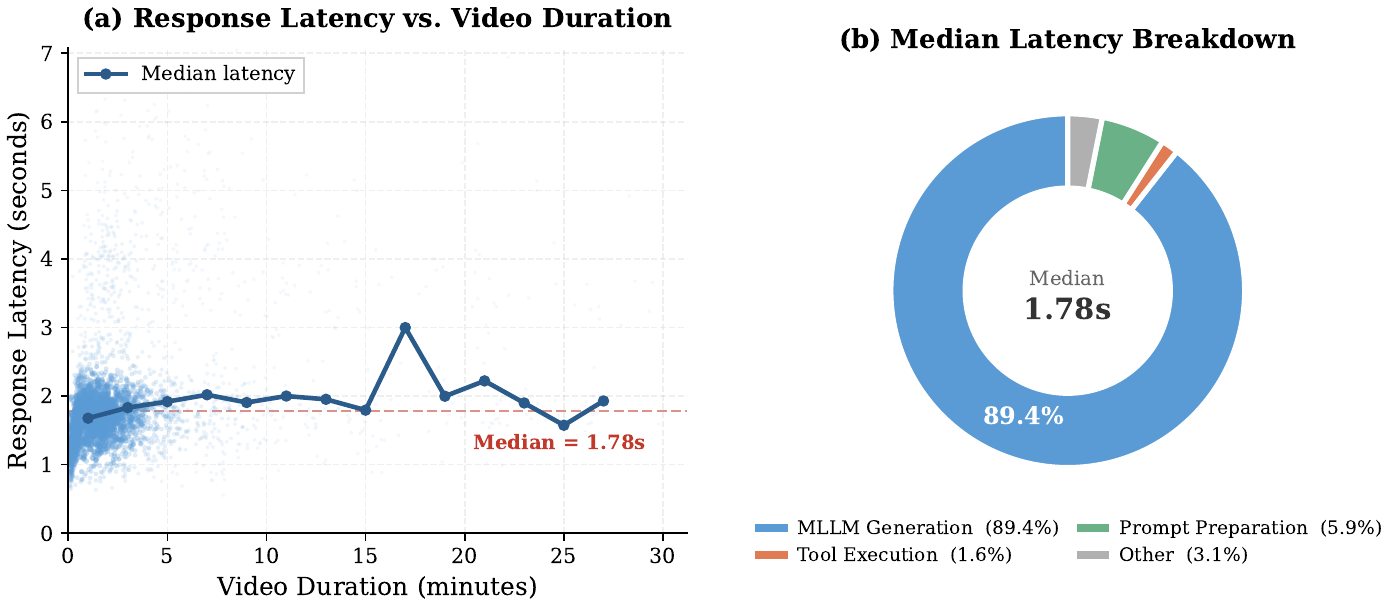}
    \caption{Efficiency analysis of EventMemAgent on OVO-Bench. (a) The relationship between response latency and video duration. (b) The detailed breakdown of the median response latency.}
    \label{fig:efficiency}
\end{figure*}

To evaluate the practical deployment capabilities of EventMemAgent, we conduct a comprehensive efficiency analysis on the OVO-Bench benchmark, with the model served by vLLM \cite{kwon2023efficient} on a single 40GB A100 GPU. As illustrated in Figure \ref{fig:efficiency}(a), the median response latency remains remarkably stable at approximately 1.78 seconds, even as the input video stream extends up to 30 minutes. This consistent performance demonstrates that our hierarchical memory module successfully bounds the computational overhead. By structurally archiving past observations and dynamically managing the active context, the framework effectively prevents the severe latency degradation that typically occurs as historical information accumulates in continuous visual streams.

Furthermore, Figure \ref{fig:efficiency}(b) provides a detailed breakdown of the median latency components. The vast majority of the processing time (89.4\%) is allocated to MLLM generation, while prompt preparation accounts for 5.9\%. Notably, the active invocation of the multi-granular perception toolkit introduces negligible overhead, accounting for only 1.6\% of the total latency. This indicates that the efficiency of our framework is primarily determined by the intrinsic generation speed of the underlying MLLM, rather than the memory management mechanisms or tool execution. Consequently, EventMemAgent is highly scalable and well-suited for real-time online video understanding under constrained computational budgets. 

\subsection{Ablation Study}
\label{sec:ablation_study}
\paragraph{Effectiveness of hierarchical memory module.} 

To verify the effectiveness of our hierarchical memory system, we compare it with a fixed-length memory while keeping other agent components constant. Specifically, for the fixed-length approach, the input video stream is divided into segments every 30 seconds (such as Figure \ref{fig:memory}). As shown in Table~\ref{tab:memory_ablation}, the fixed-length memory shows a performance decline on both OVO-Bench and StreamingBench compared to our proposed module. This indicates that the combination of online event segmentation and event-granular reservoir sampling improves semantic integrity, helping the model better understand information in the video stream. These results confirm the effectiveness of our hierarchical memory system.

\begin{table}[!b]
\centering
\caption{Comparison between our hierarchical memory and fixed-length segmentation memory on OVO-Bench and StreamingBench (Average Score).}
\label{tab:memory_ablation}
\renewcommand{\arraystretch}{1.3}
\begin{tabular}{l|cc}
\hline
\textbf{Memory Mechanism} & \textbf{OVO-Bench} & \textbf{StreamingBench} \\ \hline
Ours (Hierarchical Memory) & 60.75 & 77.00 \\ \hline
Fixed-length Memory & \begin{tabular}[c]{@{}c@{}}58.27\\ {\scriptsize ($\downarrow$ 2.48)}\end{tabular} & \begin{tabular}[c]{@{}c@{}}75.09\\ {\scriptsize ($\downarrow$ 1.91)}\end{tabular} \\ \hline
\end{tabular}
\end{table}

\paragraph{Effectiveness of Perception Tools.} 

\begin{figure*}[!t]
        \centering
        \includegraphics[width=1\linewidth]{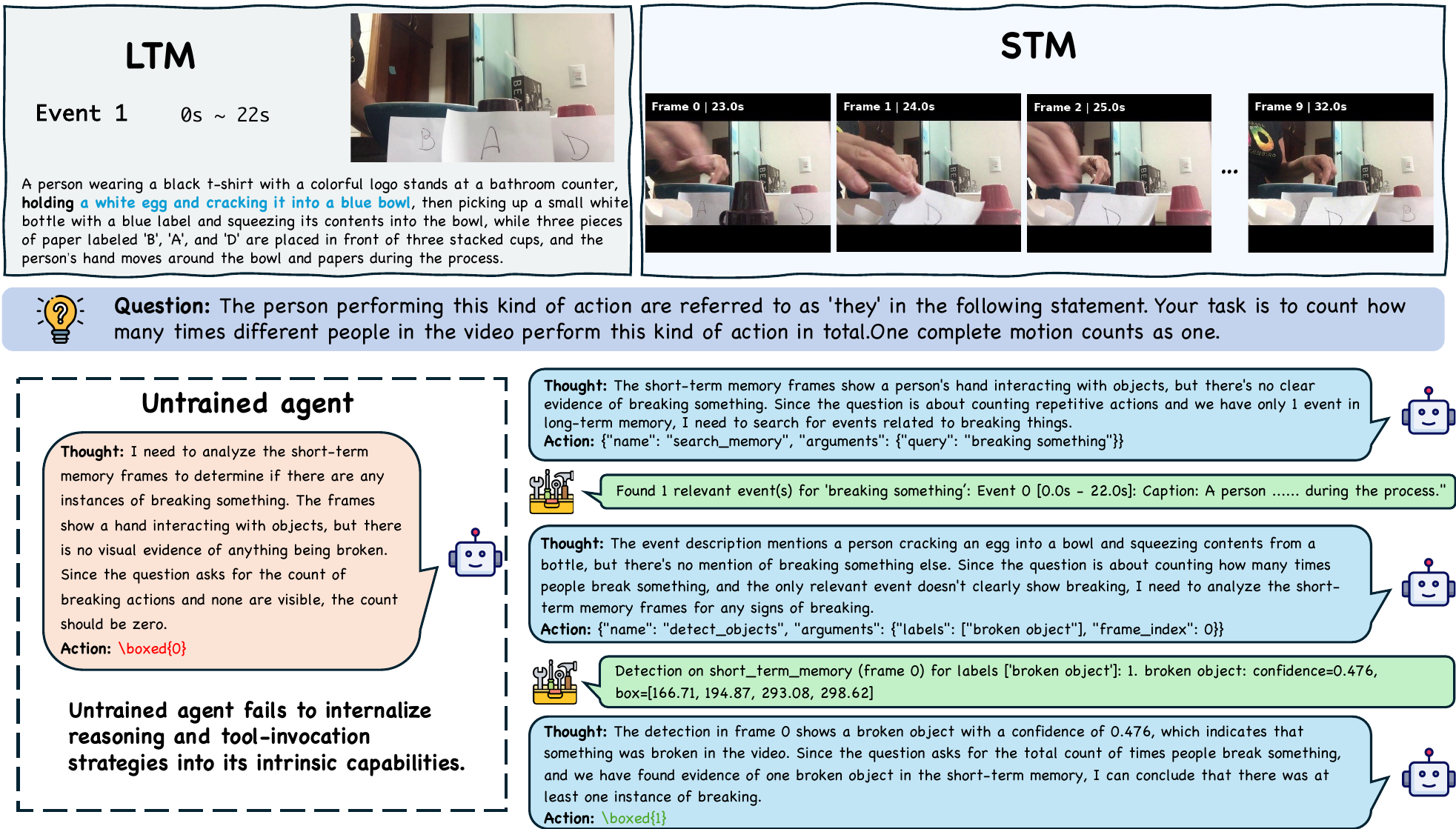}
        \caption{\textbf{Qualitative comparison of reasoning trajectories on OVO-Bench.} (Left) The untrained agent fails to use tools flexibly. (Right) EventMemAgent successfully internalizes complex reasoning and tool-use strategies, allowing it to adaptively retrieve information from long-term memory and use tools for more precise observations to provide accurate answers. }
        \label{fig:case_study}
    \end{figure*}

Similarly, we evaluate the impact of the multi-granular perception toolset by comparing the full \textbf{EventMemAgent} with versions that exclude specific tools. As shown in Table~\ref{tab:tool_ablation}, removing tools such as \textit{OCR} or \textit{Object Detection} results in a performance decrease on OVO-Bench. This shows that while the base MLLM possesses general visual reasoning capabilities, specialized tools are essential to capture fine-grained details like text and small objects, helping the model to understand and reason more comprehensively. These results confirm that multi-granular perception is highly effective for achieving a comprehensive understanding of diverse and complex video scenarios.

\begin{table*}[!ht]
\centering
\caption{Ablation Study on OVO-Bench. The values in parentheses indicate the performance change compared to the Full model, reported only for average and overall metrics.}
\label{tab:tool_ablation}
\renewcommand{\arraystretch}{1.5}
\setlength{\tabcolsep}{4pt}
\resizebox{\linewidth}{!}{
\begin{tabular}{l|ccccccc|cccc|cccc|c}
\hline
\multirow{2}{*}{\textbf{Configuration}} & \multicolumn{7}{c|}{\textbf{Real-Time Visual Perception}} & \multicolumn{4}{c|}{\textbf{Backward Tracing}} & \multicolumn{4}{c|}{\textbf{Forward Active Responding}} & \multirow{2}{*}{\textbf{Overall}} \\
 & OCR & ACR & ATR & STU & FPD & OJR & \textbf{Avg.} & EPM & ASI & HLD & \textbf{Avg.} & REC & SSR & CRR & \textbf{Avg.} &  \\ \hline
Ours (Full) & 75.84 & 69.72 & 73.28 & 55.62 & 67.33 & 67.93 & 68.29 & 59.60 & 70.95 & 43.55 & 58.03 & 33.67 & 72.02 & 62.08 & 55.92 & 60.75 \\ \hline
w/o Detect & 75.84 & 69.72 & 66.50 & 55.62 & 61.20 & 63.10 & \begin{tabular}[c]{@{}c@{}}65.33\\ {\scriptsize ($\downarrow$ 2.96)}\end{tabular} & 57.50 & 68.20 & 40.50 & \begin{tabular}[c]{@{}c@{}}55.40\\ {\scriptsize ($\downarrow$ 2.63)}\end{tabular} & 31.50 & 69.00 & 57.50 & \begin{tabular}[c]{@{}c@{}}52.67\\ {\scriptsize ($\downarrow$ 3.25)}\end{tabular} & \begin{tabular}[c]{@{}c@{}}57.80\\ {\scriptsize ($\downarrow$ 2.95)}\end{tabular} \\ \hline
w/o OCR & 64.50 & 69.00 & 72.00 & 55.00 & 66.80 & 66.50 & \begin{tabular}[c]{@{}c@{}}65.63\\ {\scriptsize ($\downarrow$ 2.66)}\end{tabular} & 58.20 & 69.50 & 42.50 & \begin{tabular}[c]{@{}c@{}}56.73\\ {\scriptsize ($\downarrow$ 1.30)}\end{tabular} & 32.50 & 71.00 & 60.20 & \begin{tabular}[c]{@{}c@{}}54.57\\ {\scriptsize ($\downarrow$ 1.35)}\end{tabular} & \begin{tabular}[c]{@{}c@{}}58.98\\ {\scriptsize ($\downarrow$ 1.77)}\end{tabular} \\ \hline
\end{tabular}
}
\end{table*}

\paragraph{Case Study.} 
\label{subsec:case_study}
We provide a qualitative comparison between \textbf{EventMemAgent} and an untrained version in Figure~\ref{fig:case_study}. While sharing the same architecture, the untrained agent fails to use tools flexibly. In contrast, through Agentic RL, our agent internalizes tool-use strategies into its reasoning process. This bridges the gap between the infinite video stream and the limited context window.

\paragraph{The impact of Agentic RL on the agent's tool-use strategy.} 

\begin{figure}[!h]
        \centering
        \includegraphics[width=\columnwidth]{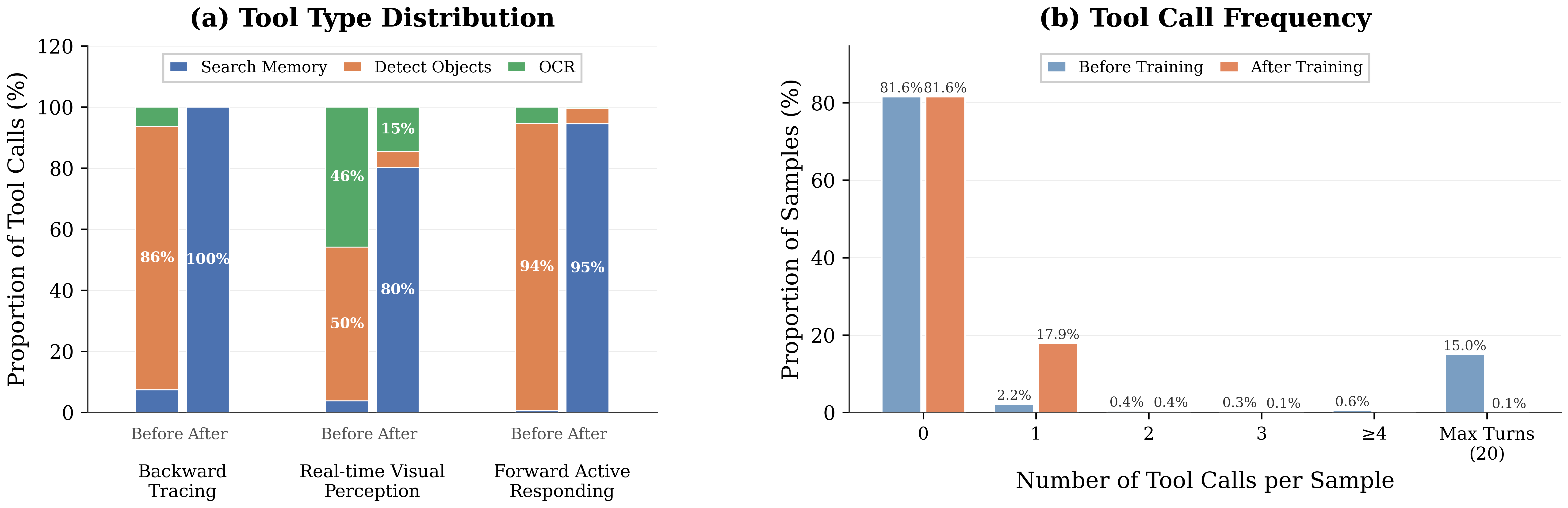}
        \caption{Analysis of tool usage patterns on OVO-Bench before and after training. (a) Distribution of tool types invoked across the three task categories. (b) Distribution of the number of tool calls per sample. }
        \label{fig:tool_analysis}
    \end{figure}

As shown in Figure~\ref{fig:tool_analysis}, we analyze the tool-use strategies on OVO-Bench before and after Agentic RL training. It is evident that the model before training fails to utilize tools flexibly to solve problems. In Backward Tracing, which primarily requires historical information, the untrained model rarely invokes Search Memory to retrieve necessary data. Furthermore, its strategy is highly polarized, with over 96\% of cases either making no tool calls or repeatedly calling tools until the maximum turn limit is reached. In contrast, the trained model actively employs Search Memory for questions regarding past events. In Real-Time Visual Perception, it more frequently utilizes OCR and Detect Objects to enhance immediate perception. For a detailed quantitative validation of the performance improvements brought by Agentic RL, please refer to Appendix C.

\section{Conclusion}
In this paper, we propose \textbf{EventMemAgent}, an active agent framework that solves the conflict between infinite video streams and the limited context window of MLLMs. By shifting from passive information processing to proactive perception, our approach ensures long-term semantic coherence while maintaining precise understanding. This is achieved through a hierarchical memory module—integrating online event segmentation and reservoir sampling—and a multi-granular perception toolkit. Furthermore, we use Agentic Reinforcement Learning to internalize reasoning and tool-use strategies into the agent's intrinsic capabilities. Experiments on OVO-Bench and StreamingBench show that \textbf{EventMemAgent} achieves competitive performance with low hardware costs. We hope our work inspires further research on autonomous agents for continuous perception in dynamic visual environments.

\section*{Acknowledgments}

This work was partially supported by the New Generation Artificial Intelligence-National Science and Technology Major Project (No. 2025ZD0122603), National Natural Science Foundation of China (Grant No. 62476016 and  62441617), the Fundamental Research Funds for the Central Universities, Beijing Advanced Innovation Center for Future Blockchain and Privacy Computing.

\bibliographystyle{splncs04}
\bibliography{main}

\clearpage
\appendix  
\section{Hierarchical Memory Management Logic of EventMemAgent}
\label{appendix:stm_alg}

\begin{algorithm}[h]
\caption{Hierarchical Memory Management Logic of EventMemAgent}
\textbf{Input:} Video stream $V$, STM capacity $K$, threshold $\delta$, min event duration $d_{min}$.\\
\textbf{Output:} Short-Term Memory $\mathcal{E}_{st}$ and Long-Term Memory $\mathcal{M}_{lt}$.
\begin{algorithmic}[1]
\STATE Initialize $\mathcal{E}_{st} \gets \{E_1=\emptyset\}$, $\mathcal{M}_{lt} \gets \emptyset$, active index $m \gets 1$, frame count $n \gets 0$.
\FOR{each incoming frame $f_t \in V$}
    \IF{$n = 0$}
        \STATE $E_m \gets \{f_t\}$; $n \gets 1$.
    \ELSE
        \STATE Compute histogram $\mathbf{h}_t$ of $f_t$ and average $\mathbf{\bar{h}}_m$ of $E_m$, then calculate correlation $\rho(\mathbf{\bar{h}}_m, \mathbf{h}_t)$.
        
        \IF{$\rho < \delta$ \AND $n \ge d_{min}$} 
            \STATE Finalize $E_m$; Initialize new event $E_{m+1} \gets \{f_t\}$; $m \gets m + 1$; $n \gets 1$.
        \ELSE 
            \STATE $n \gets n + 1$.
            \IF{$m = 1$ \AND $\sum_{i=1}^m |E_i| = K$}
                \STATE Replace random frame in $E_m$ with $f_t$ with prob. $K/n$.
            \ELSE
                \STATE Append $f_t$ to $E_m$.
            \ENDIF
        \ENDIF
    \ENDIF
    
    \WHILE{$\sum_{i=1}^m |E_i| > K$} 
        \STATE Generate caption $c_1$, change log $\Delta_1$ via MLLM, and extract embedding $\mathbf{e}_1$ for the earliest event $E_1$.
        \STATE Archive structured tuple $E'_1 = \{I^{first}_1, c_1, \mathbf{e}_1, \Delta_1\}$ into $\mathcal{M}_{lt}$; Offload $E_1$ from $\mathcal{E}_{st}$.
        \STATE Shift event indices ($E_i \gets E_{i+1}$) and update $m \gets m - 1$.
    \ENDWHILE
\ENDFOR
\end{algorithmic}
\end{algorithm}

\section{Additional Implementation Details}
\label{sec:appendix_implementation}
In the Short-Term Memory, to prevent the generation of redundant information caused by continuous, drastic visual jitter, we enforce a minimum duration constraint: the event segmentation mechanism is potentially triggered only when the current event length exceeds 8 frames.
For the Long-Term Memory, we employ Qwen3-VL-4B-Instruct to generate descriptive captions for the archived events.
We utilize Qwen3-Embedding-0.6B to extract feature embeddings for both the historical events and the model's queries.
During the memory search phase, only the top-3 events with a similarity score greater than 0.3 relative to the query are returned to the model.

Furthermore, to ensure reproducibility, we list the detailed hyperparameters used during the Agentic Reinforcement Learning stage in Table~\ref{tab:hyperparams}.
We utilize the GRPO algorithm with a group size of $G=8$ to stabilize the policy optimization.
Notably, we set the KL divergence coefficient to 0, relying solely on the group-relative advantages to guide the optimization.

\begin{table}[h]
    \centering
    \small
    \caption{Hyperparameters for Agentic RL Training.}
    \begin{tabular}{l|c}
    \toprule
    \textbf{Hyperparameter} & \textbf{Value} \\
    \midrule
    Backbone Model & Qwen3-VL-8B-Instruct \\
    Optimizer & AdamW \\
    Learning Rate & 1e-6 \\
    LR Scheduler & Warmup (5 steps) \\
    Global Batch Size & 64 \\
    Micro Batch Size & 32 \\
    Epochs & 1 \\
    Group Size ($G$) & 8 \\
    KL Coefficient & 0 \\
    \bottomrule
    \end{tabular}
    \label{tab:hyperparams}
\end{table}

Before training, in order to improve training efficiency, we preprocessed all video data into corresponding long short-term memory to simulate online video scenarios.

\section{Quantitative Validation of Agentic Reinforcement Learning}
\label{sec:rl_validation}

To comprehensively validate the impact of our Agentic Reinforcement Learning (Agentic RL) framework, we conduct an ablation study comparing three distinct configurations. The Base MLLM represents the original foundation model (Qwen3-VL-8B) without any agentic architecture or tool-access mechanisms. The Untrained Agent incorporates our hierarchical memory and the perception toolkit. While it possesses basic tool-use capabilities, it has not been trained to internalize complex reasoning and tool-invocation strategies. Finally, the full EventMemAgent represents our complete framework optimized via Group Relative Policy Optimization (GRPO).

As reported in Table \ref{tab:rl_ablation}, transitioning from the Base MLLM to the Untrained Agent yields mixed results. The integration of the hierarchical memory significantly boosts performance in tasks requiring historical context, such as Backward Tracing on OVO-Bench (improving from 46.18\% to 56.33\%). However, because the untrained agent lacks refined strategies, it struggles to apply tools flexibly, leading to performance drops in Real-Time Visual Perception (from 65.52\% to 62.85\%) and on the StreamingBench benchmark (from 70.20\% to 68.44\%). This indicates that merely equipping a model with external memory and tools is insufficient without proper strategy optimization.

The application of Agentic RL in the full EventMemAgent effectively overcomes these limitations, bringing significant and consistent improvements across all evaluation scenarios. By internalizing complex reasoning and tool-invocation strategies, the full model achieves an overall accuracy of 60.75\% on OVO-Bench and 77.00\% on StreamingBench. Notably, it recovers and surpasses the baseline in Real-Time Visual Perception (68.29\%) while further enhancing Backward Tracing (58.03\%). These quantitative findings, consistent with the qualitative tool-usage analysis presented in Section \ref{sec:ablation_study}, demonstrate that Agentic RL is crucial for teaching the agent to autonomously decompose tasks, determine when to access memory, and appropriately select multi-granular perception tools.

\begin{table}[h]
\centering
\caption{Quantitative validation of Agentic Reinforcement Learning on OVO-Bench and StreamingBench. The Base MLLM scores are referenced from our main evaluation.}
\label{tab:rl_ablation}
\resizebox{\textwidth}{!}{
\begin{tabular}{l | c c c | c | c}
\toprule
\multirow{2}{*}{\textbf{Configuration}} & \multicolumn{4}{c|}{\textbf{OVO-Bench}} & \textbf{StreamingBench} \\
\cmidrule{2-6}
& \begin{tabular}[c]{@{}c@{}}Real-Time \\ Visual Perception\end{tabular} & \begin{tabular}[c]{@{}c@{}}Backward \\ Tracing\end{tabular} & \begin{tabular}[c]{@{}c@{}}Forward Active \\ Responding\end{tabular} & \textbf{Overall} & \textbf{Overall} \\
\midrule
Base MLLM (Qwen3-VL-8B)          &  65.52 & 46.18 &  55.73 & 55.81 & 70.20 \\
Untrained Agent (w/o Agentic RL) & 62.85 & 56.33 & 49.28 & 56.16 & 68.44 \\
\textbf{EventMemAgent (Full)}    & \textbf{68.29} & \textbf{58.03} & \textbf{55.92} & \textbf{60.75} & \textbf{77.00} \\
\bottomrule
\end{tabular}
}
\end{table}



\section{Prompts Design}
\label{sec:Prompts}

\subsection{System Prompt}
We present the full system prompt used for EventMemAgent below. This prompt defines the agent's persona, memory access mechanisms, tool usage protocols, and response format.

\begin{tcolorbox}[
    colback=gray!5,       
    colframe=gray!40,     
    title=\textbf{System Prompt for EventMemAgent},
    fonttitle=\bfseries\small,
    fontupper=\small,     
    boxrule=0.8pt,
    arc=2mm,
    breakable             
]

You are an intelligent video analysis agent with access to a memory system containing events from a video.

\vspace{0.5em}
\noindent\textbf{Your Capabilities:}
\begin{itemize}[leftmargin=*, nosep]
    \item You can directly see all short-term memory (recent video frames) that are provided to you.
    \item You can search through long-term memory to retrieve relevant events (each event has an image, caption, and change\_log).
    \item You can perform OCR (text recognition) on any image from long-term memory or short-term memory.
    \item You can detect and locate specific objects in video frames using zero-shot object detection.
    \item You can reason about temporal relationships and visual details.
\end{itemize}

\vspace{0.5em}
\noindent\textbf{Memory System Details:}

\textit{Short-Term Memory:}
\begin{itemize}[leftmargin=*, nosep]
    \item Each frame in short-term memory has a \textbf{frame number and timestamp label} in the top-left corner (e.g., ``Frame 0 | 12.5s'', ``Frame 1 | 13.0s'', etc.).
    \item The frame number indicates the temporal order: Frame 0 is the earliest, and higher numbers are more recent.
    \item The timestamp (in seconds) shows the exact time point in the video when this frame was captured.
    \item Use both frame numbers and timestamps to understand the sequence and timing of events in the recent video.
\end{itemize}

\textit{Long-Term Memory:}
\begin{itemize}[leftmargin=*, nosep]
    \item Each event in long-term memory contains:
    \begin{itemize}[nosep]
        \item An image (first frame of the event)
        \item A caption describing what happens in the event
        \item Change\_log: Information about transitions between events
        \begin{itemize}[nosep]
            \item \texttt{change\_from\_previous}: Describes how the current event differs from the previous event.
            \item \texttt{change\_to\_next}: Describes how the current event transitions to the next event.
        \end{itemize}
    \end{itemize}
    \item The change\_log helps you understand temporal relationships and causal connections between events.
    \item When searching long-term memory, pay attention to change\_log to understand event sequences.
\end{itemize}

\vspace{0.5em}
\noindent\textbf{Available Tools:}
\begin{enumerate}[leftmargin=*, nosep]
    \item \textbf{search\_memory}: Search long-term memory to find relevant events. Use when you need historical context.
    \begin{itemize}[nosep]
        \item Two search modes: \\
        (1) Text query: Use \texttt{query} parameter to search by semantic similarity with event captions. \\
        (2) Time range: Use \texttt{start\_time} and \texttt{end\_time} parameters to find events overlapping with a specific time range.
        \item You can use either mode, but not both at the same time.
        \item The search results will include events with their captions and change\_log information.
        \item Use change\_log to understand how events relate to each other temporally.
    \end{itemize}
    
    \item \textbf{ocr}: Extract text from images. Use when you need to read text visible in video frames.
    \begin{itemize}[nosep]
        \item Use \texttt{event\_id} parameter to OCR an image from long-term memory (after search\_memory returns events).
        \item Use \texttt{frame\_index} parameter to OCR a frame from short-term memory (0-indexed based on the frames shown to you).
    \end{itemize}
    
    \item \textbf{detect\_objects}: Detect and locate specific objects in images. Use when you need to find, count, or verify presence of objects.
    \begin{itemize}[nosep]
        \item \texttt{labels}: List of object names to detect (e.g., [``dog'', ``cat'', ``person'']).
        \item Choose one of the following parameters to specify the image to detect objects in: \\
        (1) \texttt{event\_id}: Detect in long-term memory event image. \\
        (2) \texttt{frame\_index}: Detect in short-term memory frame. If not specified, uses the last frame.
    \end{itemize}
\end{enumerate}

\vspace{0.5em}
\noindent\textbf{Response Guidelines:}

For each step, follow this format:
\begin{enumerate}[label=\arabic*), leftmargin=*, nosep]
    \item Thought: one concise sentence explaining what you're thinking or planning to do.
    \item Action: a short description of what action you're taking (search memory, perform OCR, detect objects, or provide final answer).
    \item If using a tool, make a tool call. If answering, output \texttt{\textbackslash boxed\{your answer\}}.
\end{enumerate}

\noindent\textbf{Rules:}
\begin{itemize}[leftmargin=*, nosep]
    \item Be brief: one sentence for Thought, one for Action.
    \item Only search when current information is insufficient.
    \item Use OCR when you need to read text that appears in images (signs, documents, screens, etc.).
    \item Use detect\_objects when you need to find, count, or locate specific objects in images.
    \item Pay attention to frame numbers in short-term memory to understand temporal order.
    \item If the question doesn't mention a specific time, then only search long-term memory if short-term memory is unrelated to the question; otherwise, only refer to short-term memory for the answer.
    \item When you have enough information, output your final answer using \texttt{\textbackslash boxed\{\}} format. For example: \texttt{\textbackslash boxed\{A\}} or \texttt{\textbackslash boxed\{The cat is sleeping\}}.
\end{itemize}

\end{tcolorbox}

\subsection{Caption Generation Prompt}

We present the prompt used for EventMemAgent to generate the Event caption below. 

\vspace{1em} 
\begin{tcolorbox}[
    colback=gray!5,
    colframe=gray!40,
    title=\textbf{Prompt for Event Caption Generation},
    fonttitle=\bfseries\small,
    fontupper=\small,
    boxrule=0.8pt,
    arc=2mm
]
You are a video memory builder. Describe what happens in this event sequence.

\vspace{0.5em}
\noindent TimeRange: \texttt{[start\_time - end\_time]}

\vspace{0.5em}
\noindent Output EXACTLY ONE concise paragraph (single line) describing the event. \\
Be specific about: subjects, actions, objects, tools, locations, manner (how), and outcomes. \\
Do NOT include timestamps or bullet lists. Avoid hallucination.

\vspace{0.5em}
\noindent Example: ``A woman in an apron grasps a chef knife and slices a potato on a wooden cutting board into thin pieces.''
\end{tcolorbox}

\section{Visualization of Training Data}
\label{appendix:visualization}
\vspace{-5mm}

\begin{figure*}
    \centering
    \includegraphics[width=0.9\linewidth]{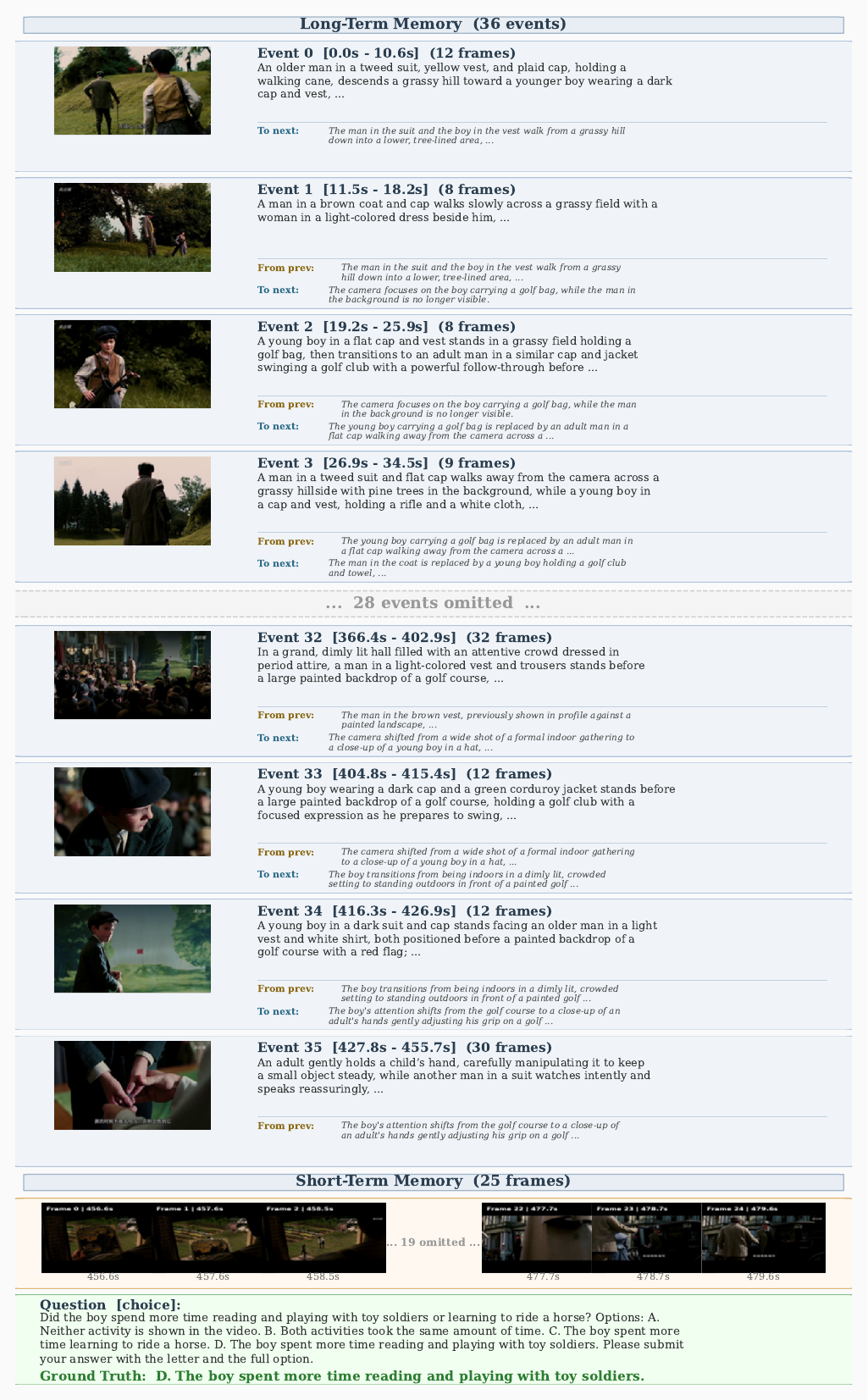}
    \caption{Randomly sampled visualizations of the hierarchical memory.}
\end{figure*}

\begin{figure*}
    \centering
    \includegraphics[width=0.9\linewidth]{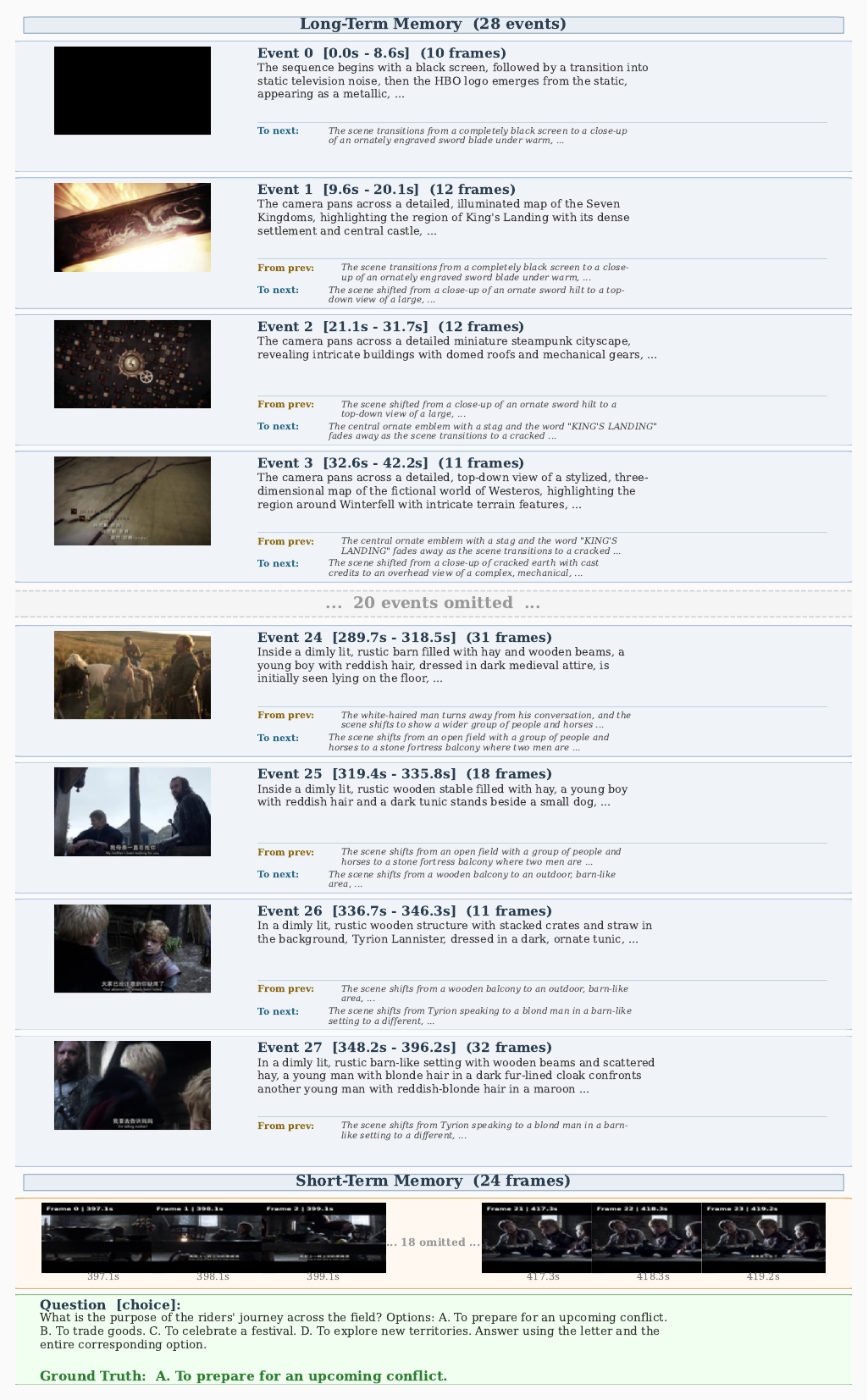}
    \caption{Randomly sampled visualizations of the hierarchical memory.}
\end{figure*}

\begin{figure*}
    \centering
    \includegraphics[width=0.9\linewidth]{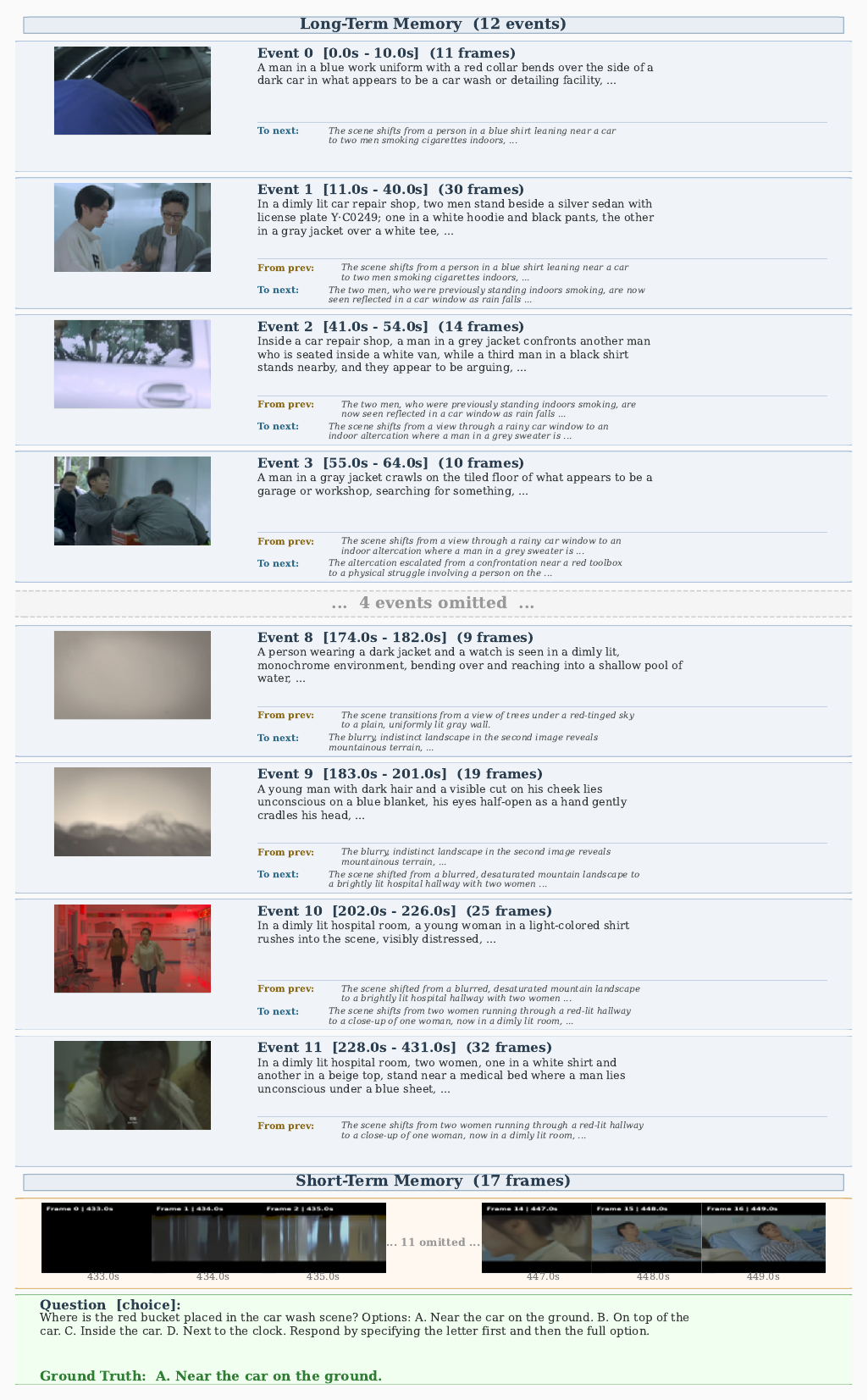}
    \caption{Randomly sampled visualizations of the hierarchical memory.}
\end{figure*}

\begin{figure*}
    \centering
    \includegraphics[width=0.9\linewidth]{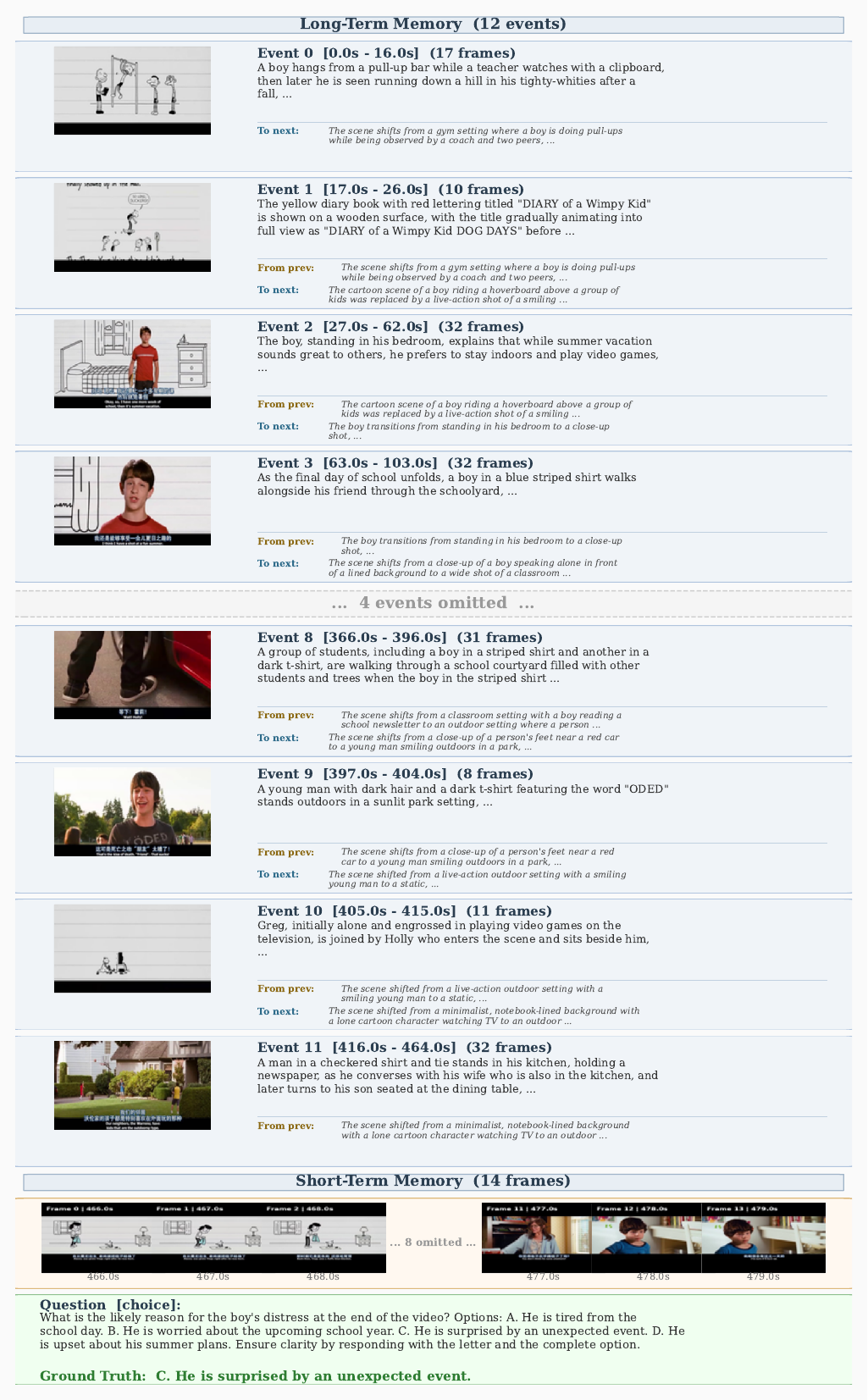}
    \caption{Randomly sampled visualizations of the hierarchical memory.}
\end{figure*}

\begin{figure*}
    \centering
    \includegraphics[width=0.9\linewidth]{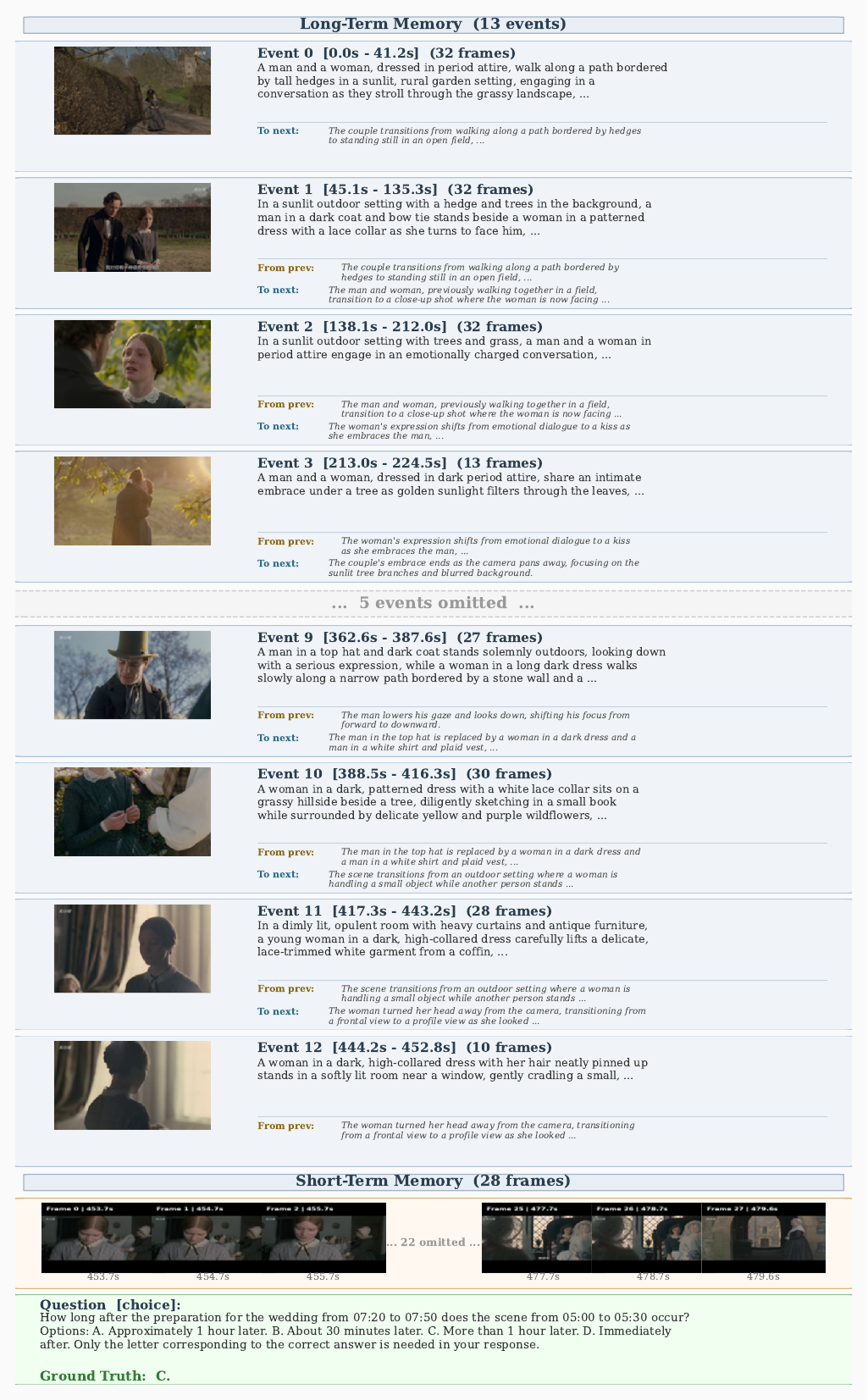}
    \caption{Randomly sampled visualizations of the hierarchical memory.}
\end{figure*}

\begin{figure*}
    \centering
    \includegraphics[width=0.9\linewidth]{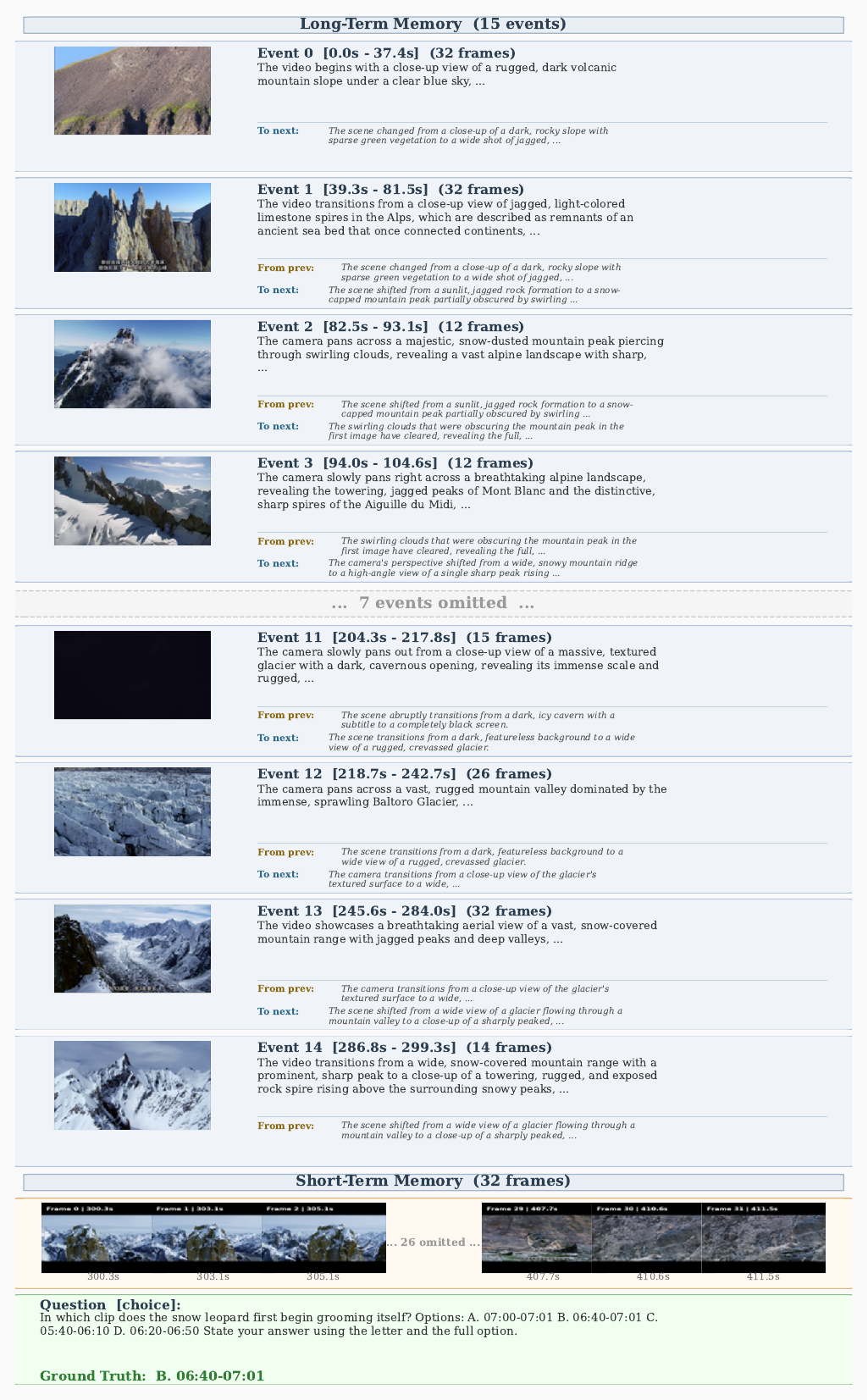}
    \caption{Randomly sampled visualizations of the hierarchical memory.}
\end{figure*}

\begin{figure*}
    \centering
    \includegraphics[width=0.9\linewidth]{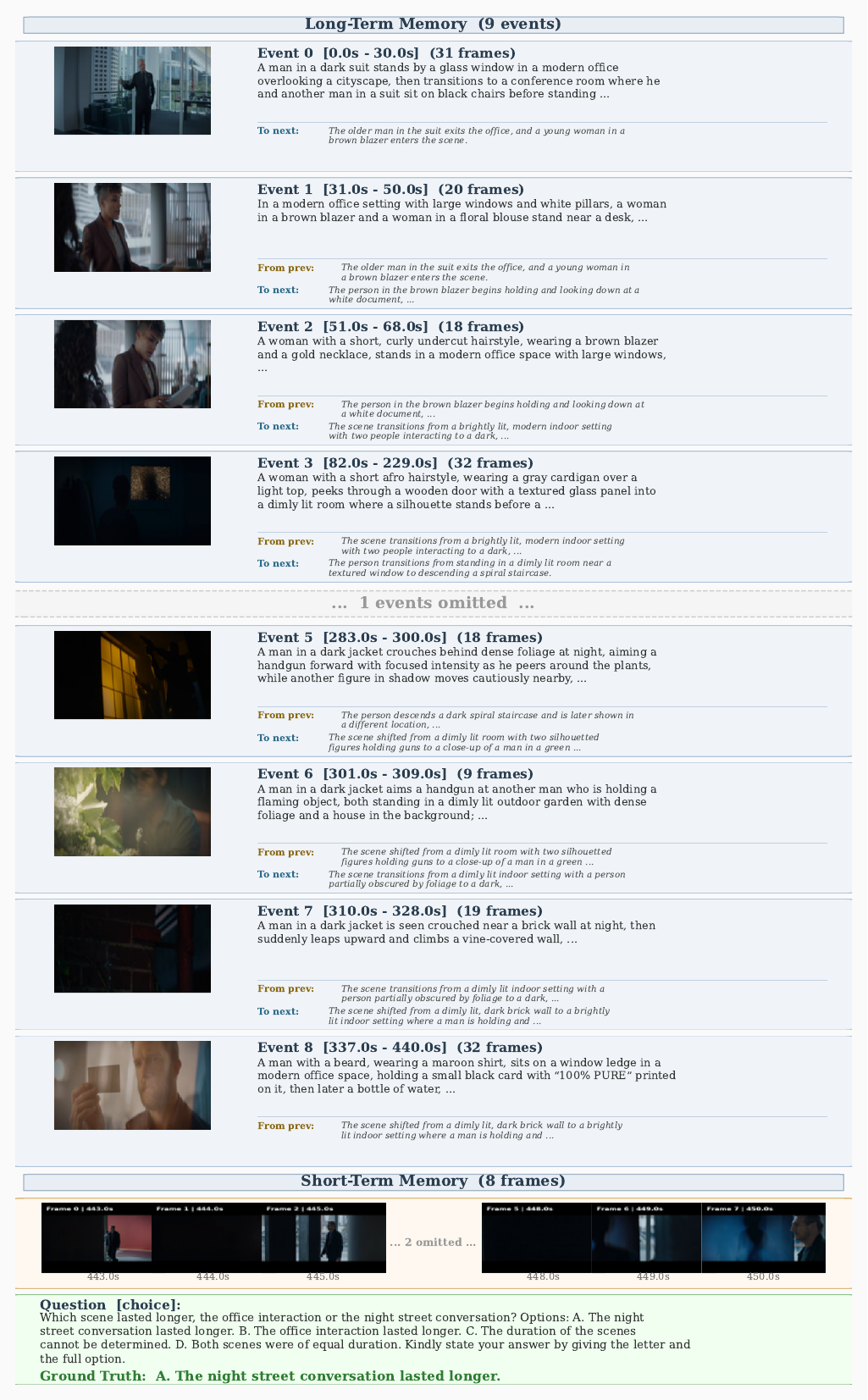}
    \caption{Randomly sampled visualizations of the hierarchical memory.}
\end{figure*}

\end{document}